%% file: 00-main.tex
\documentclass[journal]{IEEEtran}
\usepackage{ulem}
\input{header.tex}

\newif\ifarxiv
\arxivtrue

\begin{document}
\title{Homotopy-Preserving, Optimal UAV-Trajectory Generation based on Spline Subdivision}

\author{Ruiqi Ni$^1$, Teseo Schneider$^2$, Daniele Panozzo$^3$, Zherong Pan$^4$, Xifeng Gao$^1$% <-this % stops a space
\thanks{$^{1}$Ruiqi Ni and Xifeng Gao are with Department of Computer Science, Florida State University (rn19g@my.fsu.edu and gao@cs.fsu.edu). $^{2}$Teseo Schneider is with the Department of Computer Science, University of Victoria (teseo@uvic.ca).
$^{3}$Daniele Panozzo is with the Department of Computer Science, New York University (panozzo@nyu.edu). $^{4}$Zherong Pan is with the Department of Computer Science, University of Illinois Urbana-Champaign (zherong@illinois.edu).}}
\allowdisplaybreaks

\markboth{IEEE Transactions on Robotics,~Vol.~14, No.~8, August~2015}%
{Shell \MakeLowercase{\textit{et al.}}: Bare Demo of IEEEtran.cls for IEEE Journals}



\maketitle

\begin{figure*}[ht!]
\centering
\scalebox{0.9}{\includegraphics[width=\linewidth]{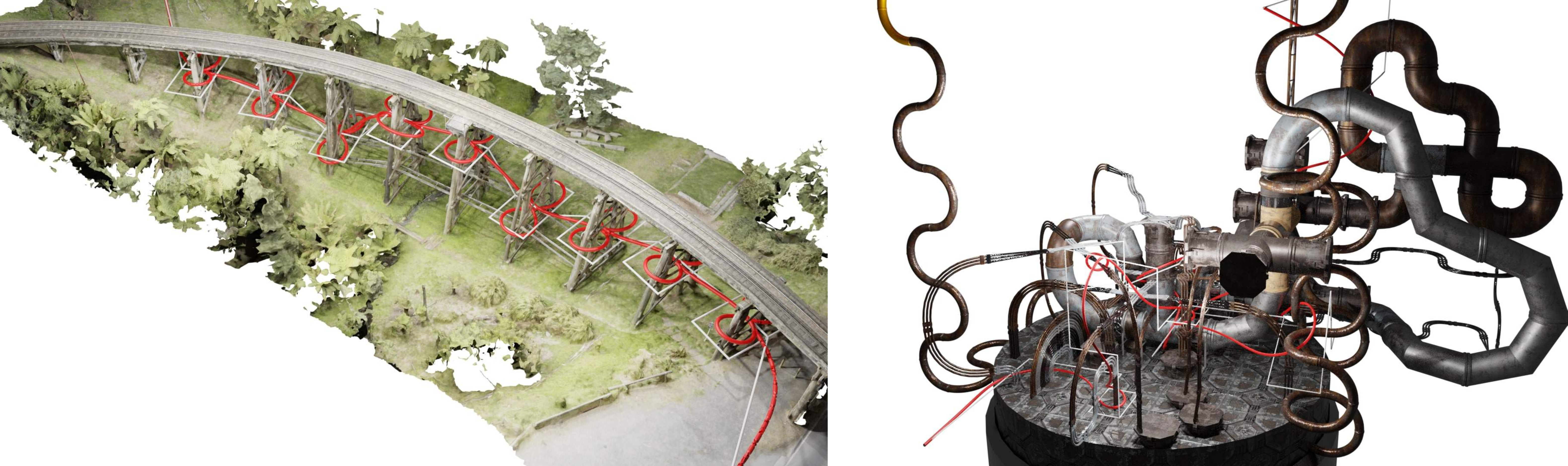}}
\caption{\small{\label{fig:teaser}Optimized (red) trajectories by our method for a bridge (left) and a machinery used in factories (right) which are challenging environments containing complex geometrical and topological features. Initial trajectories are in white color.
During the optimization, our approach maintains the safe distance clearance, the homotopy class of the trajectory, and the dynamic actuation limits, and takes 1.2hrs and 0.25hrs to finish the solve respectively. }}
\end{figure*}

\begin{abstract}
Generating locally optimal UAV-trajectories is challenging due to the non-convex constraints of collision avoidance and actuation limits. We present the first local, optimization-based UAV-trajectory generator that simultaneously guarantees the validity and asymptotic optimality for known environments. \textit{Validity:} Given a feasible initial guess, our algorithm guarantees the satisfaction of all constraints throughout the process of optimization. \textit{Asymptotic Optimality:} We use an asymptotic exact piecewise approximation of the trajectory with an automatically adjustable resolution of its discretization. The trajectory converges under refinement to the first-order stationary point of the exact non-convex programming problem. Our method has additional practical advantages including joint optimality in terms of trajectory and time-allocation, and robustness to challenging environments as demonstrated in our experiments.
\end{abstract}


% \fi
\IEEEpeerreviewmaketitle

\input{01-introduction.tex}
\input{02-related.tex}
\input{03-problem.tex}
\input{04-method.tex}
\input{05-experiment.tex}
\input{06-conclusion.tex}

\section*{Acknowledgment}

This work was supported in part by NSF-IIS-(1910486 and 1908767).

\ifCLASSOPTIONcaptionsoff
  \newpage
\fi



%\bibliographystyle{IEEEtran}
\bibliographystyle{IEEEtranS}
\bibliography{refe}
\ifarxiv
\input{07-appendixA.tex}
\input{08-appendixB.tex}
\input{09-appendixC.tex}
\fi

\end{document}

%% file: header.tex
\usepackage{amsmath,amsfonts,amssymb}
\usepackage{prettyref}
\usepackage{mathrsfs}
\usepackage{graphicx}
\usepackage{wrapfig}
\usepackage{subfig}

\usepackage{enumitem}
\usepackage{MnSymbol}
\usepackage{multirow}
\usepackage{booktabs}
\usepackage{algorithm}
\usepackage[table]{xcolor}
\usepackage[noend]{algpseudocode}
\PassOptionsToPackage{hyphens}{url}
\usepackage[colorlinks,allcolors=gray,hypertexnames=true]{hyperref} 

\newtheorem{remark}{Remark}
\newtheorem{theorem}{Theorem}[section]
\newtheorem{lemma}[theorem]{\TE{Lemma}}
\newtheorem{assume}[theorem]{\TE{Assumption}}
\newtheorem{proposition}[theorem]{\TE{Proposition}}
\newtheorem{definition}[theorem]{\TE{Definition}}
\newtheorem{corollary}[theorem]{\TE{Corollary}}

\algnewcommand{\LineComment}[1]{\State \(\triangleright\) #1}
\algdef{SE}[DOWHILE]{Do}{doWhile}{\algorithmicdo}[1]{\algorithmicwhile\ #1}
\newcommand*{\colorboxed}{}
\def\colorboxed#1#{%
  \colorboxedAux{#1}%
}
\newcommand*{\colorboxedAux}[3]{%
  \begingroup
    \colorlet{cb@saved}{.}%
    \color#1{#2}%
    \boxed{%
      \color{cb@saved}%
      #3%
    }%
  \endgroup
}
\usepackage{xcolor}
\definecolor{OrangeRed}{rgb}{0.82, 0.33, 0}
\definecolor{Green}{rgb}{0.2, 0.8, 0}
\definecolor{Blue}{rgb}{0.2, 0.2, 0.8}

\newrefformat{Fig}{Figure~\ref{#1}}
\newrefformat{fig}{Figure~\ref{#1}}
\newrefformat{par}{Section~\ref{#1}}
\newrefformat{appen}{Appendix~\ref{#1}}
\newrefformat{sec}{Section~\ref{#1}}
\newrefformat{sub}{Section~\ref{#1}}
\newrefformat{rem}{Remark~\ref{#1}}
\newrefformat{table}{Table~\ref{#1}}
\newrefformat{alg}{Algorithm~\ref{#1}}
\newrefformat{Alg}{Algorithm~\ref{#1}}
\newrefformat{Def}{Definition~\ref{#1}}
\newrefformat{Ass}{Assumption~\ref{#1}}
\newrefformat{Prop}{Proposition~\ref{#1}}
\newrefformat{Cor}{Corollary~\ref{#1}}
\newrefformat{Thm}{Theorem~\ref{#1}}
\newrefformat{Lem}{Lemma~\ref{#1}}
\newrefformat{step}{Step~\ref{#1}}
\newrefformat{ln}{Line~\ref{#1}}
\newrefformat{eq}{Equation~\ref{#1}}
\newrefformat{pb}{Problem~\ref{#1}}
\newrefformat{it}{Item~\ref{#1}}
\newrefformat{te}{Term~\ref{#1}}
\def\Eqref Eq:#1:{\eqref{eq:#1}}
\newrefformat{Eq}{Equation~\Eqref#1:}

\newcommand{\TE}[1]{\textbf{#1}}

\newcommand{\FPP}[2]{\frac{\partial{#1}}{\partial{#2}}}
\newcommand{\FPPR}[2]{{\partial{#1}}/{\partial{#2}}}
\newcommand{\FPPTR}[2]{{\partial^2{#1}}/{\partial{#2}^2}}
\newcommand{\FPPT}[2]{\frac{\partial^2{#1}}{\partial{#2}^2}}

\newcommand{\fmin}[1]{\underset{#1}{\text{min}}\;}
\newcommand{\fmax}[1]{\underset{#1}{\text{max}}\;}
\newcommand{\argmin}[1]{\underset{#1}{\text{argmin}}\;}

\newcommand{\argminP}[1]{\text{argmin}\;}
\newcommand{\ST}{\text{s.t.}\;}

%% file: 01-introduction.tex
\section{Introduction}
Unmanned Aerial Vehicle (UAV) finds many real-world applications in inspection, search and rescue, and logistic automation. A key challenge in safe UAV-trajectory planning is to find locally optimal flying strategies, in terms of energy/time efficiency and smoothness, while accounting for various dynamic and kinematic constraints. A valid trajectory needs to satisfy two constraints to be physically realizable: the actuation limits must be respected (dynamic constraint), and the robot should maintain a safety distance from the boundary of the freespace (kinematic constraint). These two constraints combined define a non-convex and non-smooth feasible domain, which is notoriously difficult to handle for optimization-based motion planners \cite{zucker2013chomp,park2012itomp,schulman2014motion}. %For example, the safe-distance constraints are typically formulated using the distance function between a line-segment on the trajectory and an edge of the environmental obstacle, but this function is non-differentiable when the line-segment is nearly parallel with the edge, thus making it unsuitable for optimization techniques relying on first and second order derivatives.

Prior works tackle the non-convex, non-smooth constraints in one of three ways: infeasible recovery, relaxation, and global search. \TE{Infeasible Recovery:} Nonlinear constrained optimizers, e.g. Sequential Quadratic Programming (SQP), have been used in \cite{zucker2013chomp,9121729,sun2020fast} to directly handle non-convex constraints. These methods rely on the constraints' directional derivatives and penalty functions to pull infeasible solutions back to the feasible domain. However, their feasibility is not guaranteed:  an example of a typical scenario of infeasible solutions involving an environment with multiple, thin-shell obstacles is discussed in \cite{schulman2014motion}. \TE{Relaxation:} In \cite{ding2018trajectory,deits2015efficient}, relaxation schemes are used to limit the solutions to a (piecewise) convex and smooth subset. For example, the freespace is approximated by the union of convex subsets and a UAV-trajectory is represented by piecewise splines, where each piece is constrained in one convex subset. These relaxed formulations can be efficiently handled by modern (mixed-integer) convex optimization tools. However, these methods can only find sub-optimal solutions because the convex subset is a strict inner-approximation with a non-vanishing gap (and increasing the approximation power increases the algorithmic running time). For environments with narrow passages, relaxation can even turn a feasible problem into an infeasible one. \TE{Global Search:} Kinodynamic-RRT$^*$ \cite{webb2012kinodynamic} and its variants can find globally optimal UAV trajectories in a discrete space and stochastic gradient descend \cite{kalakrishnan2011stomp} will approach locally optimal solutions only if the descendent direction is approximately by infinitely many samples. However, these algorithms take excessively large number of samples. To mitigate their computational cost, prior methods such as \cite{mechali2019rectified} have to terminate sampling early and then numerically rectify the solutions, and these numerical rectification methods still suffer from the limitations of infeasible recover or relaxation approaches.

\TE{Main Results:} We propose a new formulation of local, optimization-based motion planning for UAV-trajectories with robustness and asymptotic optimality guarantee. Our method represents the UAV-trajectory using smooth curves, encoded as piecewise B\'{e}zier curves. For each B\'{e}zier piece, we formulate the collision-free constraints between its control polygon and the environmental geometry. We show that these constraints can be formulated as a summation of distance functions between geometric primitives (e.g. point-triangle and edge-edge pairs). Each distance function is differentiable when restricted to the feasible domain and the restriction can be achieved using a primal interior point method (P-IPM). By using fewer geometric primitive pairs for the collision avoidance, we further propose an \textit{inexact} P-IPM algorithm that achieves a speedup up to 15$\times$, comparing to P-IPM. The convergence guarantees, under infinite precision, are provided for both the P-IPM and \textit{inexact} P-IPM methods. 

As compared with infeasible recovery methods that maintain both primal and dual variables, our primal-only formulation provides guaranteed solution feasibility throughout the entire process of optimization. This feature allows our planner to answer anytime re-planning queries as in \cite{likhachev2005anytime} because the planner can be terminated at any iteration and return a feasible solution. And compared with relaxation-based methods, our method can asymptotically minimize the sub-optimality gap due to the control-polygon approximation of the true trajectory. This is achieved by using an adaptive B\'{e}zier subdivision scheme with user-controllable termination criterion.

Our methods contribute to the solving of UAV planning problem in three ways: 1) Combined with RRT-style approaches, we can robustly optimize a feasible initial guess. 2) Starting from a feasible initial guess, our primal solver is guaranteed to preserve the homotopy class of the trajectory. This feature is useful, for example, for inspection planners where the order of inspection cannot be altered and \prettyref{fig:teaser} demonstrates that our approach can be potentially used for complex environment inspections. 3) Our formulation achieves simultaneous optimality in trajectory shapes and time allocations.%As compared with \cite{sun2020fast} where a separate decision variable is needed for time-allocation of each spline, we only need a global time re-scaling.

This paper is an extension of our previous work \cite{ruiqi2021} with 1) more technical details of the inexact P-IPM approach, 2) a complete convergence analysis showing that, within a finitely many iterations, our inexact algorithm convergences to a solution with the same quality as P-IPM, and 3) additional experiments that compare the performances of both approaches, and demonstrate the effectiveness of our approach on challenging settings.  To foster researches along this direction, we release the source code of our implementation at \href{https://github.com/ruiqini/traj-opt-subdivision}{https://github.com/ruiqini/traj-opt-subdivision}, and also make public of the dataset used for our experiments on \href{https://drive.google.com/file/d/1DM86tO0wUNef2G3BqX1U6s52vXGT5wuf/view?usp=sharing}{Google Drive}.

%% file: 02-related.tex
\section{Related Work}
Trajectory optimization for a general robotic system can be solved using dynamic programming \cite{slegers2006nonlinear}, direct collocation \cite{von1993numerical}, and path-integral control \cite{kalakrishnan2011stomp,williams2016aggressive}. While widely applicable, these methods require a search in a high-dimensional space and are thus computationally demanding and sensible to numerical failures. For the special case of UAV dynamics, a more robust and efficient approach has been proposed in \cite{5980409}, where the algorithm optimizes a reference position trajectory (global search) and then recovers the control inputs making use of differential flatness (local optimization). Our method follows the same strategy but offers the additional advantage of feasibility guarantees and locally optimality of the original non-convex optimization problem.

\TE{Global Search} provides an initial trajectory that routes the UAV from a start to a goal position and usually optimizes the trajectory by minimizing a state- or time-dependent objective function. Global search can be accomplished using sampling-based motion planner \cite{webb2013kinodynamic,li2016asymptotically}, mixed-integer programming \cite{deits2015efficient}, discrete search \cite{liu2017search}, and fast-marching method \cite{8462878}. All these methods are approximating the globally optimal UAV-trajectory under some assumptions: optimal sampling-based motion planner approaches the optimal solution after a sufficient number of samples have been drawn; mixed-integer programming and discrete search restricts the decision space to a disjoint-convex or discrete subset, respectively; fast-marching methods discretize both the trajectory and the environment on a uniform grid. Global search is essential not only for providing an initial guess to the local optimization, but also for satisfying additional constraints. Typical constraints include visitation order \cite{grundel2004formulation}, homotopy class \cite{6425970}, and coverage \cite{guerrero2013uav,di2015energy}. These constraints can be formulated into the mixed-integer programming via additional binary decision variables \cite{deits2015efficient} or into discrete search algorithms such as A$^*$ by pruning trajectories that violate the constraints \cite{liu2017search}. All these methods are orthogonal to our work and any global search method with the collision-free guarantee can be used as the initialization step of our method.

\TE{Local Optimization} complements the global search by refining the trajectories in a neighborhood of the initial guess. This step lifts the restricted search space assumption in the global search and further minimizes the objective function in the continuous decision space, until a local optimal solution is attained. In addition, the local step ensures that the solution satisfies kinematics and dynamics constraints so that the trajectory is executable on hardware. All prior works formulate the local step as a continuous or discrete optimization problem. The continuous problem can only be solved in the obstacle-free cases with a closed-form solution \cite{webb2012kinodynamic,liu2017search}. In \cite{8462878,9196789,9121729}, obstacle-free constraints are relaxed to be convex and the resulting discrete optimization is guaranteed to be solvable using primal-dual interior point method (PD-IPM). In \cite{8758904,8206214}, obstacle-free constraints are considered in its original, non-convex form, which is solved by PD-IPM but without feasibility guarantee. In this work, we show that the feasibility guarantee can be provided using P-IPM.

\TE{Time-Allocation} is an essential part of local optimization formulated as part of the dynamic constraints. Early works \cite{richter2016polynomial,8462878} first optimize the shape of the trajectory and then allocate time for each discrete segment, and the resulting trajectory is sub-optimal with respect to the time variables. It has been shown that, for a fixed trajectory, optimal time-allocation can be achieved using bang-bang solutions computed via numerical integration \cite{pham2014general}. The trajectories computed using numerical integration by \cite{pham2014general} are locally optimal in either shape variables or time variables but not both at the same time. The latest works \cite{sun2020fast,9196789} achieve shape-time joint optimality via bilevel optimization or weighted combination. We following \cite{9196789} and optimize a weighted combination of shape and time cost functions.

%% file: 03-problem.tex
\section{Problem Statement}
We propose a new method for local optimization of UAV trajectories. A UAV moves along the center-of-mass trajectory $p(t,W)$ with $t\in[0,T]$ where $T$ is the travel time and $W$ is the set of decision variables. A local optimizer takes an initial trajectory as input and locally updates it to minimize a cost function:
\begin{align}
\label{eq:obj}
\mathcal{O}(W,T)=\int_0^T c(p(t,W),t) dt + R(p(t,W),T),
\end{align}
where $c(p,t)$ is the state-dependent cost (e.g., dependent on velocity, acceleration, jerk, or snap) and $R(p,T)$ is the terminal cost. In the meantime, a feasible UAV trajectory should satisfy a set of kinematic, dynamics, and application-dependent constraints. In this paper, we assume that dynamics constraints take the form of velocity and acceleration limits:
\begin{align}
\label{eq:limits}
\|\dot{p}(t,W)\|\leq v_{\max}\quad\|\ddot{p}(t,W)\|\leq a_{\max}\quad\forall t\in[0,T],
\end{align}
which are semi-infinite constraints in the variable $t$. Here $\{v,a\}_{\max}$ are the maximum allowable velocity and acceleration. We further assume only the collision-free kinematic constraint that can take different forms depending on how obstacles are represented. Prominent environmental representations are point clouds and triangle meshes. These two representations can be uniformly denoted as a tuple of discrete elements $\mathcal{E}=\langle\mathcal{P},\mathcal{L},\mathcal{T}\rangle$ where $\mathcal{P}$ is a set of points, $\mathcal{L}$ is a set of line segments, and $\mathcal{T}$ is a set of triangles. The distance between a point $x$ and the environment is denoted as $\text{dist}(x,\mathcal{E})$. Throughout the paper, we slightly abuse notation and define $\text{dist}(\bullet,\bullet)$ as closest distance between a pair of geometric objects (e.g. points, line segments, triangles, convex hulls, point clouds, and triangle meshes). Our complete local trajectory optimization problem can be written as:
\footnotesize
\begin{equation}
\begin{aligned}
\label{eq:lopt}
\argmin{W,T}\mathcal{O}(W,T)\quad
\ST&\forall t\in[0,T]
\begin{cases}
\|\dot{p}(t,W)\|\leq v_{\max}  \\
\|\ddot{p}(t,W)\|\leq a_{\max} \\
\text{dist}(p(t,W),\mathcal{E})\geq d_0
\end{cases},
\end{aligned}
\end{equation}
\normalsize
where $d_0$ is an uncertainty-tolerating safety distance. \prettyref{eq:lopt} is among the most challenging problem instances in operations research due to the semi-infinite constraints in time $t$ and the non-smoothness of the function $\text{dist}(\bullet,\bullet)$ in general \cite{8462878,hauser2018semi}.

\subsection{Trajectory Representation}
% \begin{wrapfigure}{r}{0.2\textwidth}
% \centering
% \vspace{-10px}
% \includegraphics[width=0.2\textwidth]{composite.pdf}
% \vspace{-8px}
% \caption{\label{fig:bezier} \small{Composite $C^2$ B\'{e}zier curve with $M=5$, $w$ consists of red control points, blue points are linear interpolations of red points, and green points are linear interpolations of blue points. The UAV trajectory (dashed) is the B\'{e}zier curve with 2 red control points in the middle and one blue / green control point on each side.}}
% \vspace{-10px}
% \end{wrapfigure}
We represent $p(t,W)$ as a composite B\'{e}zier curve where each B\'{e}zier piece has a constant degree $M$. We note that this generic definition allow for curves of arbitrary continuity. The continuity across different B\'{e}zier pieces is achieved by constraining the control points. The constraints are linear and can thus be removed from the system based on St\"{a}rk's construction \cite{10.5555/581820}.

%% file: 04-method.tex
\section{Subdivision-Based P-IPM}
We propose an iterative algorithm, inspired by the recently proposed incremental potential contact handling scheme \cite{10.1145/3386569.3392425}, to find locally optimal solutions of \prettyref{eq:lopt} with guaranteed feasibility.
\begin{figure}[th]
\vspace{-5px}
\centering
\includegraphics[trim=12.5cm 0 0 0,clip,width=0.5\textwidth]{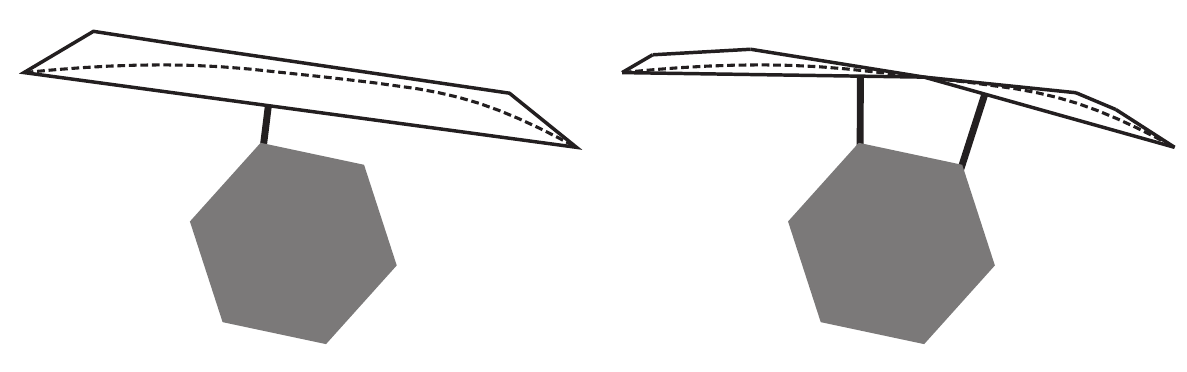}
\put(-170,10){(a)}
\put(-40,10){(b)}
\vspace{-5px}
\caption{\label{fig:comparison} \small{(a): Our method imposes distance constraints between the convex hull of control points and the obstacle. (b): Our constraints can be refined by subdivision.}}
\vspace{-12px}
\end{figure}

\subsection{Method Overview}
 The key idea of \cite{10.1145/3386569.3392425} is to use a primal-only (instead of primal-dual) interior point method to solve collision-constrained nonlinear optimization problems. This is achieved by transforming all constraints including collision avoidance constraints into log-barrier functions, and reducing the problem to non-linear, unconstrained optimization. Although primal-dual approaches have better numerical stability in many problems, their convergence relies on sufficient smoothness and Mangasarian-Fromovitz constraint qualification (MFCQ) along the central path, which do not hold for collision constraints, often leading to failures in feasibility recovery. In contrast, given a feasible initial guess, line-search based, primal-only solvers are guaranteed (even in floating point implementations), to stay inside the feasible domain. The key to establish this invariance is the use of a robust line-search scheme that prevents solution from leaving the feasible domain, in our case the major challenge is rejecting iterations with a jump in homotopy class or violation of collision-free constraints. For other constraints, namely the acceleration and velocity limits, we show that they can be handled in a similar manner (Section~\ref{sec:time-opt}). It has been shown in \cite{10.1145/3386569.3392425} that, for linearized trajectories, these safety checks can be performed using linear continuous collision detection (CCD). In this work, we extend this construction for motion planning with curved trajectories.

A safe line-search scheme ensures feasibility but does not guarantee local optimality, i.e. convergence to a first-order critical point. Assuming gradient-based optimizers are used, the additional requirement for local optimality is the objective function \prettyref{eq:lopt} being differentiable, which is challenging due to the non-smoothness of the function $\text{dist}(\bullet,\bullet)$. Although SQP algorithms with gradient-sampling \cite{curtis2012sequential} can handle piecewise-smooth constraints such as $\text{dist}(p(t,W),\mathcal{E})\geq d_0$, they degrade the deterministic convergence guarantee to a probabilistic one in the sampling limit. In \cite{10.1145/3386569.3392425,hauser2018semi}, the authors note that two objects, which could be represented by triangle meshes, $\mathcal{E}$ and $\mathcal{E}'$ are $d_0$-apart if and only if any line segment pair in $\mathcal{LL'}\triangleq\mathcal{L}\times\mathcal{L}'$ and point-triangle pair in $\mathcal{PT'}\triangleq\mathcal{P}\times\mathcal{T}'$ or $\mathcal{P'T}\triangleq\mathcal{P}'\times\mathcal{T}$ are $d_0$-apart. As a result, the log-barrier of $\text{dist}(\bullet,\bullet)$ is equivalent to the sum of log-barrier of distance functions between primitive pairs, which can be made differentiable after an arbitrarily small perturbation.

We plan to use primal-only algorithms to solve UAV trajectory's local optimization problems. To this end, the three preconditions of the primal-only method must hold. The first condition of a feasible initial guess holds trivially by using an appropriate global searcher. With a fine-enough discretization granularity of signed distance field in \cite{8462878}, or that of sampled waypoints in \cite{liu2017search}, a feasible initial guess can be computed as long as a solution exists. For the second condition, although robust, exact continuous collision predicates exist for linear triangular meshes \cite{10.1145/2185520.2185592,tang2014fast}, they have not been developed for polynomial curves. To bridge the gap, we propose to relax the exactness and resort to a conservative CCD scheme based on the subdivision of B\'{e}zier curve. For the third condition, i.e. the differentiability of the objective function, we propose to replace the trajectory with the union of convex hulls of the control polygons. These convex hulls are triangle meshes so the differentiable log-barrier between geometric primitives can be used. The accuracy of this approximation can be made arbitrarily high via adaptive subdivision.

\subsection{P-IPM Framework}
We use a column vector $w$ to represent the set of control points of a B\'{e}zier curve. For the  $i$-th piece B\'{e}zier curve of the trajectory, we denote its control points as $A_iW$, where $A_i$ is a fixed transformation matrix of decision variables $W$ into $i$-th curve's control points. A B\'{e}zier piece can be subdivided  by a linear transformation of its control points into two pieces \cite{10.5555/557058}: we denote control points of the first piece as $D_1w$ and the control points of the second piece as $D_2w$, where $D_{1,2}$ are the fixed subdivision stencils constructed using the De-Casteljau's algorithm. Our adaptive subdivision scheme (\prettyref{alg:tsubd}) will recursively apply these stencils until a stopping criterion is met. And we keep a subdivision history $\mathcal{H}$ to maintain the subdivision transformation matrices for each subdivided B\'ezier curve piece, where $\mathcal{H}$ will be initialized with $\{A_i|i=1,...,N\}$. Our algorithm handles two meshes $\mathcal{E}$ and $\mathcal{E}'$. We use $\mathcal{E}$ to denote the mesh of the environment and $\mathcal{E}'$ denotes the mesh discretizing the UAV trajectory. After the subdivision, $\mathcal{E}'$ will be updated so that it is the union of convex hulls of the new set of control points. For each subdivided B\'ezier curve piece with control points $w$, we denote $P(w)$ as the set of $M+1$ vertices of the convex hull, $L(w)$ as the set of $(M+1)M/2$ edges connecting all pair of vertices of $P(w)$, and $T(w)$ as the set of $(M+1)M(M-1)/6$ triangles connecting any 3 vertices. As illustrated in \prettyref{fig:comparison}, given the environment $\mathcal{E}$ (represented using either a point cloud or a triangle mesh) and $\mathcal{E}'$, we can define our log-barrier function as:
\small
\begin{align}
\label{eq:log}
B(W,\mathcal{E})\triangleq
\sum_{A\in\mathcal{H}}\biggl[&\sum_{l\in\mathcal{L}, l'\in L(w)}\text{clog}(\text{dist}(l,l')-d_0)+  \\
&\sum_{p\in\mathcal{P}, t'\in T(w)}\text{clog}(\text{dist}(p,t')-d_0)+\nonumber  \\
&\sum_{t\in\mathcal{T}, p'\in P(w)}\text{clog}(\text{dist}(t,p')-d_0)\biggr]\bigg|_{w=AW},\nonumber
\end{align}
\normalsize
where $\text{dist}$ is the mollified, differentiable distance between a segment-segment or line-triangle pair, and $\text{clog}$ is the clamped log-barrier function:
\begin{align*}
\text{clog}(x)=
\begin{cases}
-\frac{(x-x_0)^2}{x}\text{log}(\frac{x}{x_0})&0<x\leq x_0  \\
0&\text{elsewhere},
\end{cases}
\end{align*}
with $x_0$ being the activation range. Note that we use a slightly modified version of $\text{clog}$ from the original paper \cite{10.1145/3386569.3392425}. Our modification preserves the second-order continuity but is essential for the correctness of our inexact algorithm introduced in Section \ref{sec:inexact}. Although there are many terms in \prettyref{eq:log}, all the terms with $\text{dist}(\bullet,\bullet)$ larger than $d_0+x_0$ are zero. Therefore, we use a bounding volume hierarchy structure to efficiently prune all the zero terms.

\begin{algorithm}[ht]
\small{\caption{$\text{NeedSub}(w)$}
\label{alg:needsub}
\begin{algorithmic}[1]
\State Return $\text{dist}(\text{Hull}(w),\mathcal{E})<d_0+x_0
\land \text{diameter}(\text{Hull}(w))>\epsilon$
\end{algorithmic}}
\end{algorithm}
%\vspace{-15px}

\begin{algorithm}[ht]
\small{\caption{$\text{Sub}(w,A,\mathcal{H})$}
\label{alg:subd}
\begin{algorithmic}[1]
%\LineComment We denote the $i$th piece of $p(t,w)$ as $A_i(s)w$
\State $<\mathcal{P}',\mathcal{L}',\mathcal{T}'>\gets<\emptyset,\emptyset,\emptyset>$
\If{$\text{NeedSub}(w) $}
\State $\mathcal{H}\gets\mathcal{H}\backslash\{A\}\bigcup\{D_1A\}\bigcup\{D_2A\}$
\State $<\mathcal{P}',\mathcal{L}',\mathcal{T}'>\gets
<\mathcal{P}',\mathcal{L}',\mathcal{T}'>\bigcup\text{Sub}(D_1w,D_1A,\mathcal{H})$
\State $<\mathcal{P}',\mathcal{L}',\mathcal{T}'>\gets
<\mathcal{P}',\mathcal{L}',\mathcal{T}'>\bigcup\text{Sub}(D_2w,D_2A,\mathcal{H})$
\Else
\State $<\mathcal{P}',\mathcal{L}',\mathcal{T}'>\gets
<P(w),L(w),T(w)>$
\EndIf
\State Return $<\mathcal{P}',\mathcal{L}',\mathcal{T}'>$
\end{algorithmic}}
\end{algorithm}
%\vspace{-15px}

\begin{algorithm}[ht]
\small{\caption{$\text{TrajSub}(p(t,W),\mathcal{H})$}
\label{alg:tsubd}
\begin{algorithmic}[1]
\State $<\mathcal{P}',\mathcal{L}',\mathcal{T}'>\gets<\emptyset,\emptyset,\emptyset>$
\For{$A\in \mathcal{H}$}
\State $<\mathcal{P}',\mathcal{L}',\mathcal{T}'>\gets<\mathcal{P}',\mathcal{L}',\mathcal{T}'>\bigcup\text{Sub}(AW,A,\mathcal{H})$
\EndFor
\State Return $\mathcal{E}'\gets<\mathcal{P}',\mathcal{L}',\mathcal{T}'>$
\end{algorithmic}}
\end{algorithm}
%\vspace{-15px}

\begin{algorithm}[ht]
\small{\caption{$\text{P-IPM}$}
\label{alg:pipm}
\begin{algorithmic}[1]
\Require $W_0,T_0$
\State $\mathcal{H}\gets\{A_i|i=1,...,N\},W\gets W_0,T\gets T_0$
\While{True}
\State $\mathcal{E}'\gets\text{TrajSub}(p(t,W),\mathcal{H})$
\State Calculate $\nabla_{W,T} O,\nabla_{W,T}^2 O$
\State Calculate $d\gets -\text{SPD}(\nabla_{W,T}^2 O)^{-1}\nabla_{W,T} O$
\State $<W,T>\gets\text{Search}(<W,T>,d)$
\If{$\|\nabla_{W,T} O\|_\infty\leq\epsilon_g$}
\State Return $<W,T>$
\EndIf
\EndWhile
\end{algorithmic}}
\end{algorithm}
%\vspace{5px}

We are now ready to summarize our main algorithm of P-IPM in \prettyref{alg:pipm}, which is a Newton-type method applied to the following unconstrained optimization problem:
\begin{equation}
\begin{aligned}
\label{eq:opt}
&\argmin{W}O(W,T) + \lambda B(W,\mathcal{E}),
\end{aligned}
\end{equation}
where $\lambda$ is the relaxation coefficient controlling the exactness of constraint satisfaction. If the cost function is differentiable, the Hessian has bounded eigenvalues, then P-IPM converges to a first-order critical point of \prettyref{eq:opt}. Finally, the function $\text{SPD}(\bullet)$ adjusts the Hessian matrix by clamping the negative eigenvalues to a small positive constant through a SVD to ensure positive-definiteness.

\subsection{Convergence Guarantee}
The convergence guarantee of \prettyref{alg:pipm} relies on the correct implementation of two functions: $\text{Search}$ (\prettyref{alg:search}) and $\text{NeedSub}$ (\prettyref{alg:needsub}). The search function updates $W$ to $W'$ and ensures that the Wolfe's first condition:
\begin{align*}
&O(W',T)+\lambda B(W',\mathcal{E})\leq O(W,T) + \lambda B(W,\mathcal{E}) + \\
&c\alpha\nabla_W\left[O(W,T) + \lambda B(W,\mathcal{E})\right]^Td,
\end{align*}
holds and $p(t,W)$ is homotopically equivalent to $p(t,W')$, where $c\in(0,1)$ is some positive constant. Instead of using CCD which requires numerically stable polynomial root finding, we use a more conservative check between the convex hull of geometric primitives before and after the update from $W$ to $W'$. This procedure is outlined in \prettyref{alg:search}. 

The $\text{NeedSub}$ function guarantees that the solution of \prettyref{eq:opt} asymptotically converge to that of \prettyref{eq:lopt}. For P-IPM, whenever there is an active log-barrier term in \prettyref{eq:log}, the convex hull of some B\'{e}zier curve with control points $w$ is at most $d_0+x_0$ away from $\mathcal{E}$. This implies we can bound the maximal distance between any curve's point and $\mathcal{E}$ as $d_0+x_0+\Delta(w)$, where $\Delta(w)$ is the diameter of the convex hull of the control polygon or any upper bound between a curve's point and its convex hull. A simple strategy that guarantees $\epsilon$-optimality is to always subdivide when 1) $\Delta(w)>\epsilon$ and 2) the distance between the convex hull of $w$ and $\mathcal{E}$ is smaller than $d_0+x_0$, where $\epsilon$ is a user-provided optimality threshold. Note that condition 2) inherently induces an adaptive subdivision scheme where all the sub-trajectories that are sufficiently faraway from $\mathcal{E}$ are not subdivided to save computation. In theory, under assumption \prettyref{Ass:A1} and $\epsilon_g=0$, we show in Section \prettyref{sec:pipm2sip} that P-IPM will converge to the local optima of the original semi-infinite oracle (\prettyref{eq:lopt}) as the coefficients of log-barrier functions ($\lambda$) and the clamp range of $\text{clog}$ functions ($x_0$) tends to zero. In practice, we show that P-IPM will terminate after a finite number of iterations under the following assumption:
\begin{assume}
\label{Ass:A1}
The objective function $\mathcal{O}$ is twice-differentiable and the sequence of $w$ generated by iterations of P-IPM is uniformly bounded.
\end{assume}
\begin{proposition}
\label{Prop:ExactTerminate}
Suppose \prettyref{Ass:A1} holds and $\epsilon_g>0$, then gradient descend method using line search \prettyref{alg:search} terminates within finitely many iterations.
\end{proposition}
\begin{IEEEproof}
See \prettyref{sec:pipm-proof2}. 
\end{IEEEproof}

\begin{algorithm}[ht]
\small{
\caption{$\text{Search}(<W,T>,d)$}
\label{alg:search}
\begin{algorithmic}[1]
\Require $\gamma,c_1\in(0,1)$
\State $\alpha\gets1, <W',T'>\gets <W,T>+\alpha d$
\While{True}
\State Safe$\gets$True
\For{$A\in \mathcal{H}$}
\If{$\text{dist}(\text{Hull}(AW\bigcup AW'),\mathcal{E})<d_0$}
\label{ln:safeguard}
\State Safe$\gets$False
\EndIf
\EndFor
\If{Safe $\land$ Wolfe's first condition}
\State Return $<W',T'>$
\Else
\State $\alpha\gets\gamma\alpha, <W',T'>\gets <W,T>+\alpha d$
\EndIf
\EndWhile
\end{algorithmic}}
\end{algorithm}

\subsection{Inexact P-IPM}
\label{sec:inexact}
A practical problem with \prettyref{alg:pipm} is that we have to add all the $M+1$ vertices $(M+1)M/2$ edges in $L(w)$ and all the $(M+1)M(M-1)/6$ triangles in $T(w)$ of each convex hull of the control polygon to \prettyref{eq:log}. Even with a bounding volume hierarchy, summing up so many terms is still time-consuming. To overcome this issue, we propose an inexact, yet preserving the guarantees of local optimality, version of P-IPM. For every B\'ezier piece, we only keep all $M+1$ points in $P(w)$ and one edge that connects the starting and ending control points of the curve's control polygon. As compared with the exact counterpart, the inexact version significantly reduces the cost of log-barrier function evaluation by a factor of $O(M^2)$ since we drop the entire triangle set $T(w)$.

Note that most edges and all triangles are omitted in constructing \prettyref{eq:log} but not in the $\text{Search}$ function to ensure feasibility. This naive inexact P-IPM is not guaranteed to converge to the local minimum of \prettyref{eq:opt} because \prettyref{alg:search} might not find a positive $\alpha$ satisfying the first Wolfe's condition due to the conservative collision check. To ensure convergence to a local minimum, a simple strategy is to keep subdividing whenever the line-search fails. By slight modifications to the line search \prettyref{alg:search}, in the following, we present an inexact P-IPM algorithm that is guaranteed to converged to the KKT point after a finite number of subdivisions.

For simplicity of presentation, we assume there is only one B\'ezier curve piece, $A(s)w$ with $s\in[0,1]$, i.e. $N=1$. After inexact P-IPM performs a series of subdivisions, the $[0,1]$ segment will be divided into consecutive intervals:
\begin{align*}
0\leq s_1<s_2<\cdots<s_{S-1}<s_S=1.
\end{align*}
Using this notation, we can rewrite the inexact log-barrier function as:
\begin{align*}
\tilde{B}(w,\mathcal{E})\triangleq\sum_{i=1}^{S-1}(s_{i+1}-s_i)
\text{clog}(\text{dist}(m_{i,i+1}'(w),\mathcal{E})-d_0),
\end{align*}
% \begin{align}
% \tilde{B}(w,\mathcal{E})\triangleq\sum_{i=1}^{S-1}(s_{i+1}-s_i)
% \biggl[&\sum_{l\in\mathcal{L}}\text{clog}(\text{dist}(l,l'(w))-d_0)+\nonumber\\
% &\sum_{p\in\mathcal{P}, p'\in P(w)}\text{clog}(\text{dist}(p, p')-d_0)+\nonumber  \\
% &\sum_{t\in\mathcal{T}, p'\in P(w)}\text{clog}(\text{dist}(t,p')-d_0)\biggr],\nonumber
% \end{align}\xifeng{check}
where we use $m_{i,i+1}'(w)$ to denote the line segment connecting the start and the end control points of curve piece with $s\in [s_i, s_{i+1}]$. Note that inexact log-barrier function is scaled by the range of each sub-curve $s_{i+1}-s_i$, which is unnecessary in its exact counterpart.

Two modifications to \prettyref{alg:search} are needed to ensure finite termination. First, we need to use a more conservative safeguard of the following form:
\begin{align*}
\text{Safe}\gets 
\text{dist}(\text{Hull}(\mathcal{P}'(w),\mathcal{P}'(w')),\mathcal{E})>
d_0+\psi(s_{i+1}-s_i),
\end{align*}
where $\psi(s)$ is some positive function for which we choose $\psi(s)=\epsilon_ss^\eta$, where $\epsilon_s,\eta$ are small positive constants. Second, we subdivide all the curve pieces that do not satisfy the safeguard condition if $\alpha$ is smaller than some $\epsilon_\alpha$. The modified line-search scheme is outlined in \prettyref{alg:searchInexact}. We also propose to save computation by reducing $\epsilon_\alpha$ after each subdivision.

\begin{algorithm}[ht]
\small{
\caption{Subdivide-Unsafe($w,w'$)}
\label{alg:subdUnsafe}
\begin{algorithmic}[1]
\For{$i=1,\cdots,S-1$}
\LineComment{Consider subdivided B\'ezier curve piece $s_i,s_{i+1}$}
\If{$\text{dist}(\text{Hull}(\mathcal{P}'(w),\mathcal{P}'(w')),\mathcal{E})<d_0+\psi(s_{i+1}-s_i)$}
\State Insert $\frac{s_i+s_{i+1}}{2}$ between $s_i$ and $s_{i+1}$
\EndIf
\EndFor
\end{algorithmic}
}
\end{algorithm}

\begin{algorithm}[ht]
\footnotesize{
\caption{$\text{Inexact-Search}(<w,T>,d,\epsilon_\alpha)$}
\label{alg:searchInexact}
\begin{algorithmic}[1]
\Require $\gamma,c_1\in(0,1),\epsilon_s>0,\eta\in(0,1/3)$
\State $\alpha\gets 1, <w',T'>\gets <w,T>+\alpha d$
\While{True}
\State Safe$\gets$True
\For{$i=1,\cdots,S-1$}
\LineComment{Consider subdivided B\'ezier curve piece $s_i,s_{i+1}$}
\If{$\text{dist}(\text{Hull}(\mathcal{P}'(w),\mathcal{P}'(w')),\mathcal{E})<d_0+\psi(s_{i+1}-s_i)$}
\State Safe$\gets$False
\EndIf
\EndFor
\label{ln:safeguardInexact}
\If{Safe $\land$ Wolfe's first condition}
\State Return $<w',T'>,\epsilon_\alpha$
\ElsIf{Safe=False $\land\alpha<\epsilon_\alpha$}
\State Subdivide-Unsafe($w,w'$)
\State $\epsilon_\alpha\gets\gamma\epsilon_\alpha$
\Else
\State $\alpha\gets\gamma\alpha, <w',T'>\gets <w,T>+\alpha d$
\EndIf
\EndWhile
\end{algorithmic}
}
\end{algorithm}

\begin{algorithm}[ht]
\small{
\caption{$\text{Inexact-P-IPM}$}
\label{alg:pipmInexact}
\begin{algorithmic}[1]
\Require $w_0,T_0,\epsilon_\alpha>0$
\State $\mathcal{H}\gets\emptyset,w\gets w_0,T\gets T_0$
\While{True}
\State $\mathcal{E}'\gets\text{TrajSub}(p(t,w),\mathcal{H})$
\State Calculate $\nabla_{w,T} O,\nabla_{w,T}^2 O$
\State Calculate $d\gets -\text{SPD}(\nabla_{w,T}^2 O)^{-1}\nabla_{w,T} O$
\State $<w,T>,\epsilon_\alpha\gets\text{Search}(<w,T>,d,\epsilon_\alpha)$
\If{$\|\nabla_{w,T} O\|_\infty\leq\epsilon_g$}
\State Return $<w,T>$
\EndIf
\EndWhile
\end{algorithmic}
}
\end{algorithm}
In summary, our inexact algorithm solves:
\begin{equation}
\begin{aligned}
\label{eq:optInexact}
&\argmin{w}O(w,T) + \lambda \tilde{B}(w,\mathcal{E}),
\end{aligned}
\end{equation}
using Newton's method \prettyref{alg:pipmInexact} with \prettyref{alg:searchInexact} being the line-search scheme. Our algorithm is guaranteed to terminate within a finite number of subdivisions as summarized below:
\begin{proposition}
\label{Props:pipmInexact}
Assuming \prettyref{Ass:A1} and $\epsilon_g>0$, then \prettyref{alg:pipmInexact} using line search \prettyref{alg:searchInexact} terminates within finitely many iterations.
\end{proposition}
\begin{IEEEproof}
See \prettyref{sec:inpipm}. 
\end{IEEEproof}
The crucial observation behind the proof is that, after infinite uniform subdivisions, the barrier function $\tilde{B}$ would approach the following integral of $\text{clog}$ in the sense of Riemann sum:
\begin{align*}
\hat{B}(w,\mathcal{E})\triangleq\int_0^1\text{clog}(\text{dist}(A(s)w,\mathcal{E})-d_0)ds,
\end{align*}
and the finite termination can be proved by bounding the error between $\tilde{B}$ and $\hat{B}$.

\subsection{Time-Optimality}\label{sec:time-opt}
\begin{figure}
\centering
%\vspace{-10px}
\begin{tabular}{cc}
\includegraphics[trim=10cm 12cm 15cm 12cm,clip,width=0.22\textwidth]{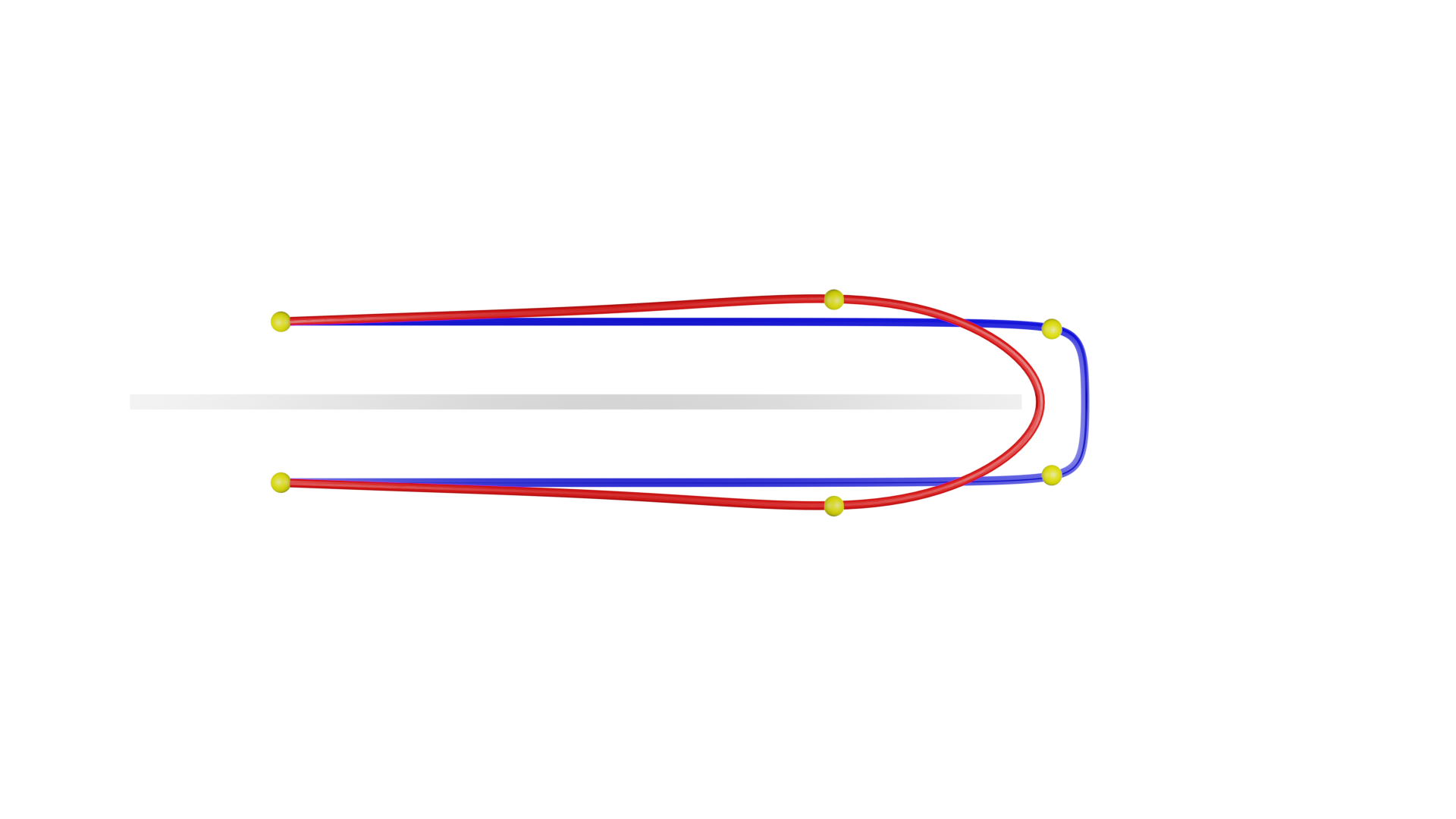} &
\includegraphics[trim=10cm 12cm 15cm 12cm,clip,width=0.22\textwidth]{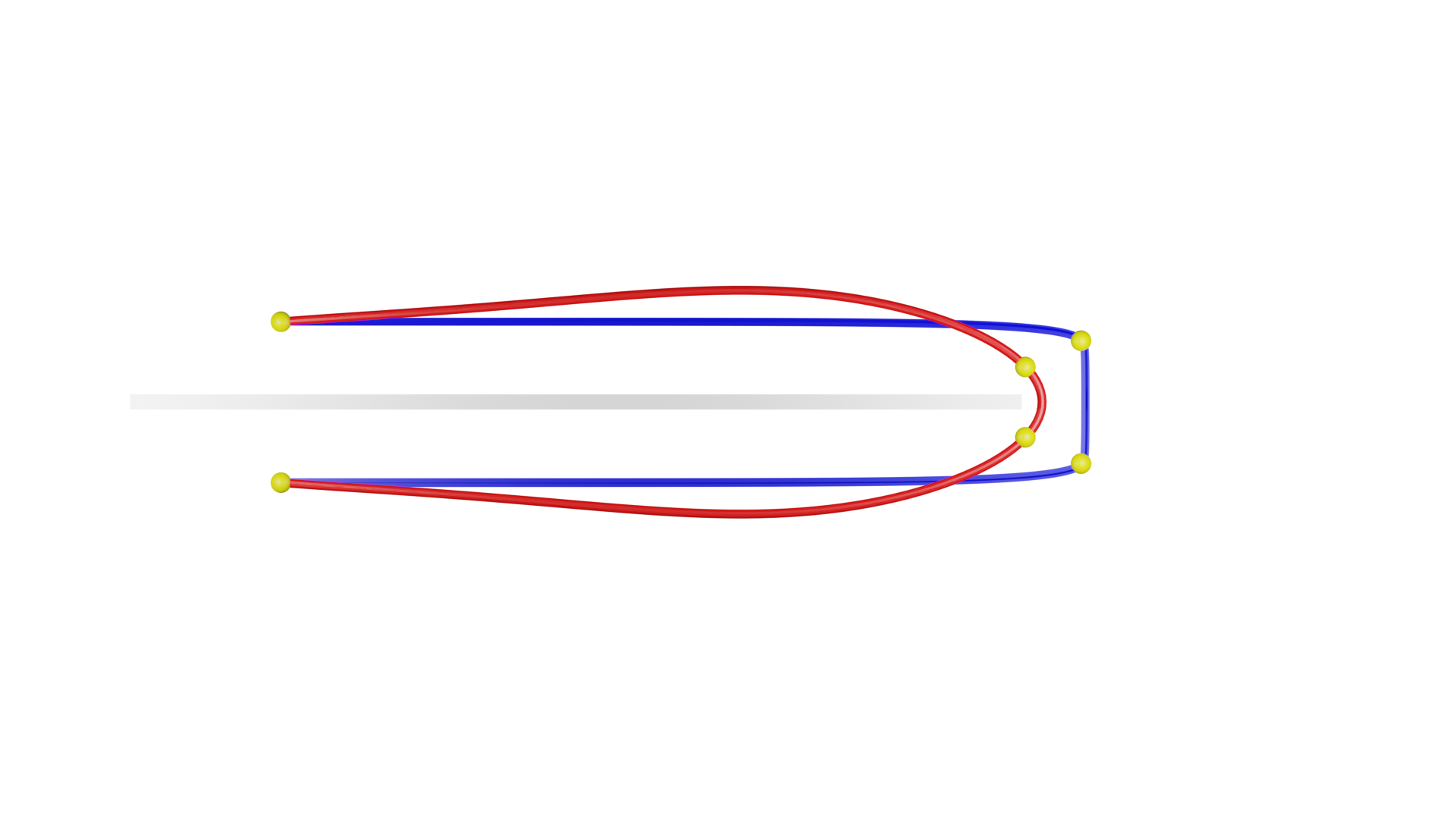}
\end{tabular}
\vspace{-4px}
\caption{\label{fig:time} \small{We optimize a curve involving a sharp turn in the middle using two different time-allocations. Left: Each curve is assigned the same amount of time. Right: The middle curve is assigned less time. For both cases, our optimizer generates similar trajectories %by adjusting the relative curve length
(yellow dots are the end points of B\'{e}zier curves).}}
\vspace{-20px}
% \end{wrapfigure}
\end{figure}

Time efficacy can be formulated as a terminal cost such as $R(p,T)=T$, combining the semi-infinite velocity and acceleration bounds. Unlike \cite{sun2020fast,8462878} that warp each B\'{e}zier curve piece using a separate time parameter, we use a single, global time re-scaling. In other words, the UAV flies through each polynomial piece for $T/N$ seconds. Although we use a single variable $T$ as compared with $N$ variables for each piece in \cite{sun2020fast,8462878}, we argue that time-optimality will not be sacrificed because our formulation allows a larger solution space for the relative length of the $N$ B\'{e}zier curves. In prior work \cite{8462878} for example, each B\'{e}zier curve is constrained to a separate convex subset of the freespace, which restricts the relative length of neighboring B\'{e}zier curves. By comparison, our method does not rely on these constraints and allow the relative length to change arbitrarily. As illustrated in \prettyref{fig:time}, we optimize two trajectories using different, fixed time allocations, and our method converges to almost identical solutions by changing the relative length of curves, which justifies the redundancy of curve-wise time variables.
% \begin{figure}[ht]
% \vspace{-10px}
% \centering
% \scalebox{.5}{\includegraphics[trim=5cm 12cm 15cm 12cm, width=\columnwidth]{fig/fig3_1.png}}
% \scalebox{.5}{\includegraphics[trim=5cm 12cm 15cm 12cm, width=\columnwidth]{fig/fig3_2.png}}
% \caption{\label{fig:time} We optimize a curve involving a sharp turn in the middle using two very different time-allocations. Top: Each curve is assigned the same amount of time. Bottom: The middle curve is assigned less time. In either case, the optimizer will return similar trajectories by adjusting the relative curve length (yellow dots are the end points of B\'{e}zier curves).}
% \vspace{-5px}
% \end{figure}

% \begin{wrapfigure}{r}{0.2\textwidth}

We use the same principle as collision constraints to handle velocity and acceleration limits, by introducing a new set of log-barrier functions. For a subdivided curve piece with control points $w$, its first and second derivatives are two new B\'{e}zier curves, where their control points are $S_1w$ and $S_2w$, and $S_{1,2}$ are the fixed transformation matrix constructed by B\'{e}zier derivative. The convex hull of their control polygons and corresponding point, edge, triangle set are denoted as $\langle P(S_1w),L(S_1w),T(S_1w)\rangle$ for velocity and $\langle P(S_2w),L(S_2w),T(S_2w)\rangle$ for acceleration. We can now define our log-barrier function approximating the velocity and acceleration limits as:
\footnotesize
\begin{align}
\label{eq:lo}
B_T(W,T)=
\sum_{i=1}^N\biggl[&\sum_{p_1\in P(S_1w)}\text{clog}(v_{max}-\|p_1\|_2/(T/N))+  \\
&\sum_{p_2\in P(S_2w)}\text{clog}(a_{max}-\|p_2\|_2/(T^2/N^2))\biggr]\bigg|_{w=A_iW}.\nonumber
%&\sum_{\dot{p}\in\dot{P}}\text{clog}((Tv_{max}/N)^2-\|\dot{p}(w)\|_2^2)+  \\
%&\sum_{\ddot{p}\in\ddot{P}}\text{clog}((T^2a_{max}/N^2)^2-\|\ddot{p}(w)\|_2^2).
\end{align}
\normalsize
P-IPM now minimizes $O(W,T) + \lambda B(W,\mathcal{E}) + \lambda B_T(W,T)$ to achieve joint optimality in terms of trajectory shape and time. Due to the convex hull property of B\'{e}zier curves, the finite value of $B_T(W,T)$ implies that B\'{e}zier curve with control points $S_1w$ is bounded by $Tv_{max}/N$ and B\'{e}zier curve with control points $S_2w$ is bounded by $T^2a_{max}/N^2$ and subdivision can make the approximation arbitrarily exact. If desired, a separate subdivision rule can be used to control the exactness, e.g. always subdivide when $\Delta(S_1w)>\dot{\epsilon}$ or $\Delta(S_2w)>\ddot{\epsilon}$. Finally, note that we need a feasible initial guess for $T$ and a sufficiently large $T$ is always feasible.

%% file: 05-experiment.tex
\section{Experiments\label{sec:experiments}}

\begin{table*}[ht!]
\centering
%\footnotesize{
\rowcolors{2}{gray!25}{white}
\setlength{\tabcolsep}{2pt}
\begin{tabular}{l|c|ccc|ccc|ccc|ccc}
\toprule
\rowcolor{gray!50}
Scene &Size
& 
$L^\text{\cite{8462878}}$ & $T^\text{\cite{8462878}}$ &  $C^\text{\cite{8462878}}$ & 
$L^\text{\cite{8206214}}$ & $T^\text{\cite{8206214}}$ &  $C^\text{\cite{8206214}}$ & 
$L^{P-IPM}$ &$T^{P-IPM}$ &  $C^{P-IPM}$& 
$L^{IP-IPM}$ &$T^{IP-IPM}$ &  $C^{IP-IPM}$\\
\midrule
 1 & 1.4M/2.7M & 70.5 & 144.3 & 168.5 &
 \textcolor{red}{82.6} & 
    \textcolor{red}{73.3} &  
    \textcolor{red}{0.5} & 
  56.3 & 30.5 & 3.4K&
 56.2 & 30.4 & 218.9\\ 
     2 & 31K/58K& 25.7 & 29.1  & 8.3  & 
              \textcolor{red}{21.5} & \textcolor{red}{26.9}  & \textcolor{red}{0.2}  
              &  16.1 & 9.6 & 13.0& 
              16.1 & 9.6 & 2.9\\      
     3 & 36K/68K& 18.7  &  24.1 &  2.5 & 
             \textcolor{red}{20.1}  & \textcolor{red}{19.0}  & \textcolor{red}{0.6} &  16.7 & 9.8 & 5.4 & 
             16.7  & 9.8  & 2.3\\  
     4 & 17K/34K& 25.1 &  24.1 & 4.0  & 
              22.8  & 23.6  &  0.1 &  19.8 & 11.3 & 15.0& 
              19.8  & 11.3  & 9.0\\  
     5 & 0.4M/0 & 49.8  &  65.0 & 6.4 & 
           \textcolor{red}{50.0}  &
           \textcolor{red}{50.0} &
           \textcolor{red}{0.3} &  39.0 & 21.7 & 400.4& 
           38.2  &  21.3 & 42.1\\      
     6 & 0.2M/0.3M & - & -  &  - &  
              \textcolor{red}{14.8} & \textcolor{red}{18.0}  & \textcolor{red}{0.2} &  14.3 & 9.7 & 176.0&
              13.8  & 9.2  &  14.9 \\    
     7 & 81K/16K & 23.2  & 38.5  &  3.8&
             \textcolor{red}{30.9}  &  \textcolor{red}{26.3} &  \textcolor{red}{0.2} &  21.6 & 12.4 & 12.6&  
             21.6 & 12.4  &6.2\\     
     8 &36K/68K& 24.5 & 23.5  &3.3  &
              \textcolor{red}{28.1} & \textcolor{red}{26.1} &\textcolor{red}{0.1}  &  18.5 & 10.7 & 13.0 &
              18.5 & 10.7  &2.2\\

%KDOP experiment result
    %   1  &  56.6 & 30.6 & 223.5\\ 
    %   2  &  16.1 & 9.6 & 3.6\\
    %   3  &  16.7 & 9.7 & 3.1\\
    %   4  &  19.8 & 11.3 & 12.6\\   
    %   5  &  39.1 & 21.7 & 63.5\\   
    %   6  &  13.8 & 9.2 & 12.0\\       
    %   7  &  21.6 & 12.4 & 6.6\\       
    %   8  &  18.4 & 10.7 & 3.3\\   
 %KDOP may be slow, because of more subdivision.

 %Exact experiment result
    %   1  &  56.3 & 30.5 & 3.4K\\ 
    %   2  &  16.1 & 9.6 & 13.0\\
    %   3  &  16.7 & 9.8 & 5.4\\
    %   4  &  19.8 & 11.3 & 15.0\\   
    %   5  &  39.0 & 21.7 & 400.4\\   
    %   6  &  14.3 & 9.7 & 176.0\\       
    %   7  &  21.6 & 12.4 & 12.6\\       
    %   8  &  18.5 & 10.7 & 13.0\\   
    
%Complex result
%bridge & 0.4M/0.7M & 72.6 & 66.9 & 4.4K
%pipe & 0.3M/0.5M & 69.4 & 45.8 & 922.5
%wall9_2 & 40.6 & 36.7 & 156.1
%wall9_20 & 41.1 & 11.6 & 111.1
 
\bottomrule
\end{tabular}
\caption{\label{table:fullres} \small{We profile the performance of \cite{8462878}, \cite{8206214}, our P-IPM method, and our Inexact P-IPM method in terms of trajectory length $L$, arrival time $T$, and computational cost $C$. The size of each scene, in terms of its numbers of vertices and faces, is listed in the form of $\#V/\#F$. $-$ denotes a failure in the provided program. Red color is used to highlight trajectories where the UAV collides with the scene.}}%use time duplicates the symbol in table
%\vspace{-5px}
%}
\end{table*}

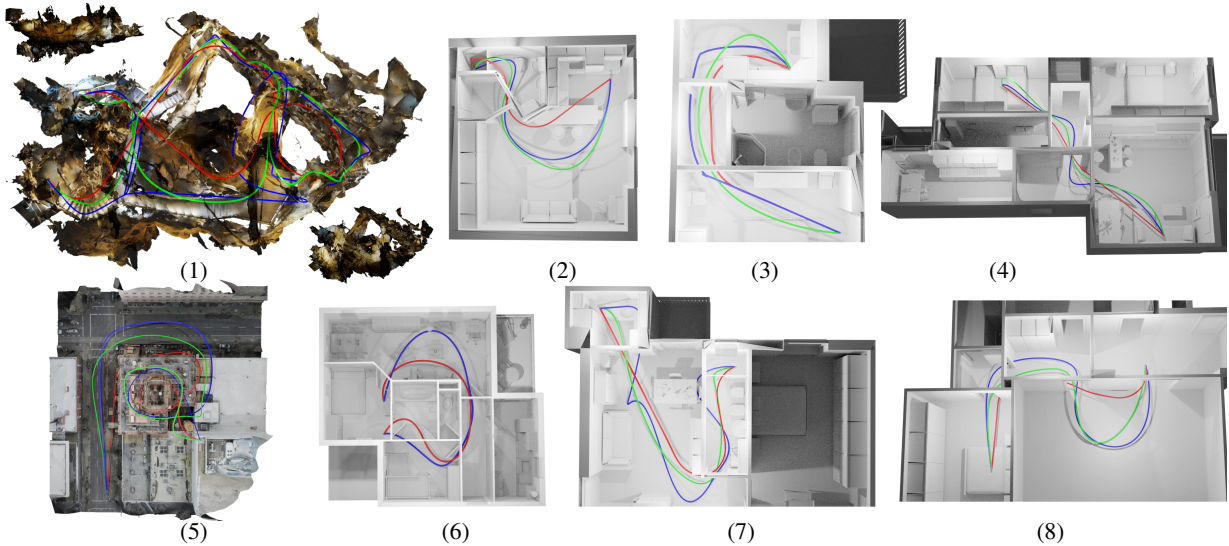
\begin{figure*}[ht!]
\vspace{-5px}
\centering
\scalebox{0.9}{
\includegraphics[trim=0 1cm 0 2cm,width=\textwidth]{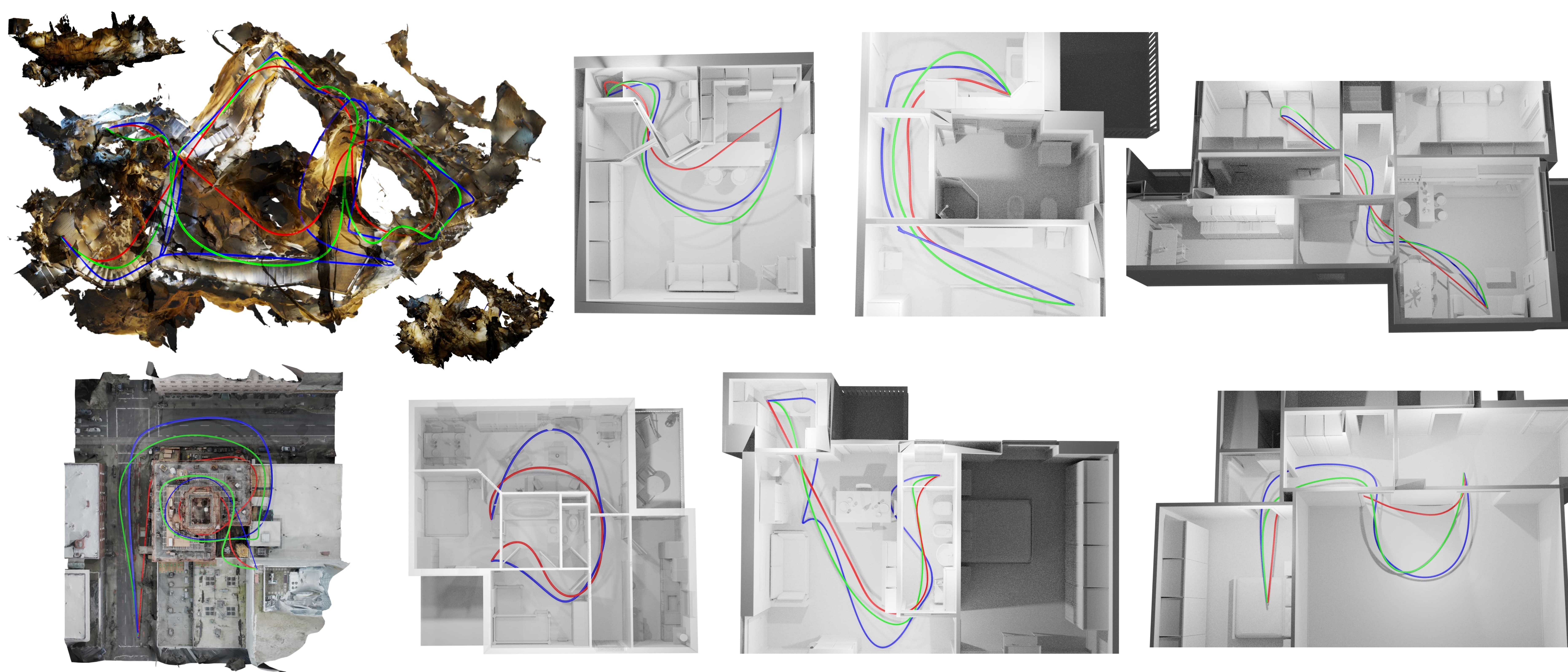}
\put(-440,100){(1)}
\put(-285,100){(2)}
\put(-200,100){(3)}
\put(-100,100){(4)}
\put(-440,-10){(5)}
\put(-330,-10){(6)}
\put(-210,-10){(7)}
\put(-80,-10){(8)}}
%\vspace{-5px}
\caption{\label{fig:compare}\small{For the scenes listed in \prettyref{table:fullres}, we compare UAV trajectories computed using our method (red), \cite{8206214} (blue), and \cite{8462878} (green) from the same initial guess. Both our method and \cite{8462878} can generate feasible trajectories and our method finds a smaller objective function (e.g. smoother trajectory and smaller travel time as shown in \prettyref{table:fullres}), while \cite{8206214} penetrates most of the environments.}}
%\vspace{-12px}
\end{figure*}

Our implementation uses C++11 and all results are computed using a single thread on a workstation with a 3.5 GHz Intel Core i9 processor. Our algorithm has the following parameters: $\lambda$ of the collision avoidance barrier term, $x_0$ of the activation range, $d_0$ of the clearance distance, $v_{\max}$, $a_{\max}$, $\epsilon$ of the subdivision threshold and $\epsilon_g$ of the stopping criterion. We use $\lambda = 10$, $x_0 = 0.1$, $d_0 =0.1$, $v_{\max} = 2.0m/s$, $a_{\max}= 2.0m/s^2$, $\epsilon = 0.1$, $\epsilon_g = 10^{-3}$ for all experiments. To match the energy setting of the works we compare with, we use the composite B{\'e}zier where each piece is degree 8 and the continuity between adjacent pieces is $C^2$, and set $c$ in the objective function as jerk energy. We use improved GJK method \cite{montanari2017improving} for convex hull collision detection.

\subsection{Comparisons}
We compare the two versions of our approach with the state-of-the-art gradient based method \cite{8206214} and corridor based method \cite{8462878} on a set of scenes, represented by either point clouds or triangle meshes, that are publicly downloadable from SketchFab \cite{sketchfab}. We use the public implementations of both methods and tune our implementation to match the parameter settings of their methods, such as the orders of curves representing the trajectory, the energy term to measuring the smoothness of the trajectory (i.e. snap, jerk), $v_{\max}$, and $a_{\max}$. The implementation of \cite{8462878} and \cite{8206214} only handles point cloud data. Therefore, we use densely sampled point clouds from the triangular meshes representing our scenes. Because of the complexity of the scenes listed in \prettyref{table:fullres}, we manually tune waypoints to ensure that there will be a valid initial trajectory generated by the global search for all the methods, i.e. \cite{8462878}, \cite{8206214}, and ours. 

\begin{figure*}[th!]
\centering
\includegraphics[width=0.9\linewidth]{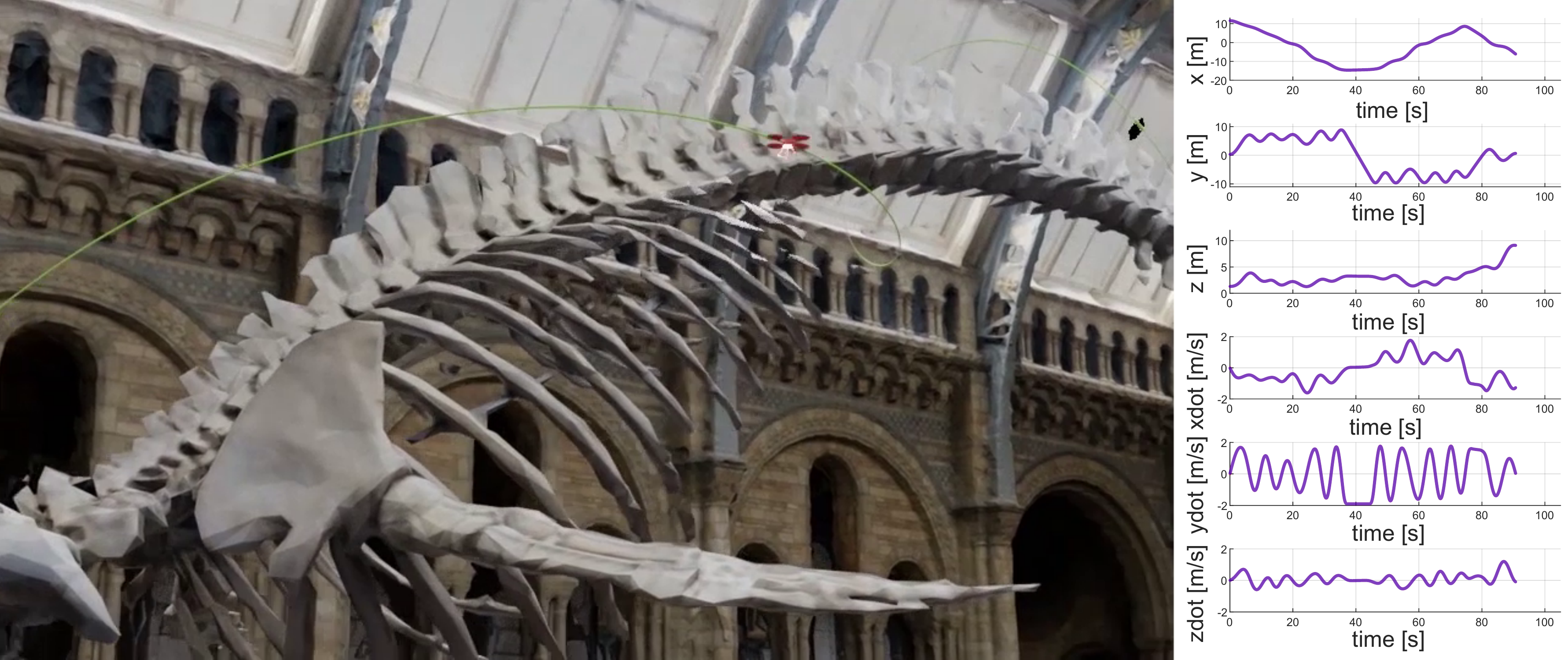}
\caption{\label{fig:museum}\small{Left: a simulation screenshot of the UAV (red) flies along a simulated trajectory that has neglectable differences from our optimized trajectory (green) in a museum. Right: each plot shows the overlaid positions (x/y/z[m]) and the speeds (xdot/ydot/zdot[m/s]) of UAV for both the simulated and our trajectories.}}
%\vspace{-15px}
\end{figure*}

We shown several benchmarks in \prettyref{table:fullres}, where most environments are represented using triangle meshes except for the 5th scene, which is a dense point cloud. While both the P-IPM and Inexact P-IPM methods produce trajectories with approximately the same lengths and travel times, the Inexact P-IPM method computes much faster than the P-IPM approach, with speedup ranging from 1.6$\times$ to 15.5$\times$. If not specifically mentioned, we perform the comparisons and simulations using only the Inexact P-IPM method since it provides similarly the robustness and convergence guarantees.

As demonstrated in \prettyref{table:fullres}, our trajectories have the highest quality and far outperform the results of \cite{8462878,8206214}. Visual results are shown in \prettyref{fig:compare}. Note that, since \cite{8206214} doesn't have a collision-free guarantee even with a valid initial trajectory, their generated trajectories often penetrate the environments as highlighted in red in \prettyref{table:fullres}. On the contrary, our method not only ensures a collision-free trajectory at every step during the optimization, but also guarantees that the trajectory is never closer to the obstacles than a user-specified safe distance. Although \cite{8462878} requires only waypoints as input, we have to modify, in a trial-and-fail way, either the waypoints or the grid resolution used by their approach so that their method can find a collision-free initialization to ensure that their trajectory optimization proceeds.

Note that, while our method obtained competitive computing efficiency to \cite{8462878} for some scenes,  \cite{8206214,8462878} typically achieve real-time computations and good quality trajectories for simple environments. However they struggle as the environments gets more complex. \cite{8206214} tends to generate invalid trajectories with collisions between the UAV and the environment. \cite{8462878} instead has a quickly growing runtime, since they have to reduce the failure rate of their initial trajectory generation by increasing the grid resolution, resulting in a similar efficiency as ours.

\subsection{Simulated UAV Flight in Challenging Environments}
In the attached video (a screenshot is shown in \prettyref{fig:museum}), we simulate the UAV flight trajectory in the challenging environment of a museum with a lot of narrow passages. The simulator is setup using \cite{5569026} on the hardware platform of a CrazyFlies nano-quadrotor. In the museum environment (represented using a dense triangle mesh with 1 million faces) we show that our optimization easily supports the generation of trajectories passing exactly through user-specified waypoints. As demonstrated in the video, our optimization can generate a high quality trajectory that has unnoticeable differences from the simulated path, allowing smooth UAV flying even during sharp turns.

%, the gradient-based method transforming collision-free constraints into soft penalty functions \cite{8206214} fails in finding feasible solutions. On the other hand, the convex relaxation method \cite{8462878} finds  a feasible solution, but their solution is far from optimal.

%% file: 06-conclusion.tex
\section{Conclusion \& Limitations}
We propose a new approach of local UAV-trajectory optimization with guaranteed feasibility and asymptotic convergence to the semi-infinite problem (\prettyref{eq:lopt}). The key to our success is the primal-only optimizer, P-IPM, equipped with an adaptive subdivision scheme for line search and constraint refinement. The line search scheme preserves the homotopy class and the constraint refinement ensures that our solution approaches that the of semi-infinite oracle with sufficiently small error. Using the same framework, we show that our method can be extended to achieve time-optimality. We further present an inexact P-IPM approach that achieves up to 10$\times$ speedup than its exact counterpart. We also present the convergence analysis showing that both exact and inexact P-IPM algorithms terminate after a finite number of iterations, and they converge to a solution of the same quality, i.e., a solution with sufficiently small gradient norm.

Our method has several limitations, leading to several avenues of future research. First, our method requires a feasible (collision-free) initial guess, which can be computed using sampling-based planners, e.g., RRT$^*$\cite{webb2012kinodynamic}. Second, our method requires the environment to be represented geometrically, which is a common assumption of all optimization-based UAV trajectory planner. We have shown in \prettyref{sec:experiments} that our method can handle several different forms of geometric representations, such as point clouds and triangle meshes, without preprocessing. Finally, although our inexact P-IPM is already up to one order of magnitude faster than the exact baseline, our method is still offline. In the future, we plan to explore randomized or first-order numerical algorithms for further speedup.

%% file: 07-appendixA.tex
\section{Convergence of Exact P-IPM to SIP\label{sec:pipm2sip}}
In this section, we analyze the convergence of \prettyref{eq:opt} to a KKT point of SIP \prettyref{eq:lopt}. For simplification, we ignore the velocity and acceleration limits, the case with these limits is similar. Here, we consider one piece B\'{e}zier trajectory, so $w$ is the same as $W$. We denote row vector of Bernstein basis as $A(s)$ with $s\in[0,1]$, then the B\'{e}zier curve is $A(s)w$.

\subsection{Perturbed Optimality\label{subsec:popt}}
We first provide a definition of (inexact) KKT point for the SIP problem following \cite{mordukhovich2013constraint}. We rewrite \prettyref{eq:lopt} as a smooth SIP by augmenting the index set:
\small
\begin{align}
\label{eq:SIP}
\argmin{w}O(w)\quad
\ST&\forall t\in[0,T], q\in\mathcal{E}\quad \text{dist}(p(t),q)\geq d_0,
\end{align}
\normalsize
which is equivalent to \prettyref{eq:lopt} (without the velocity and acceleration limits). $\text{dist}(p(t),q)$ is the distance between two points, which is a smooth function when $p(t)\neq q$. We abbreviate $O(w,T)$ as $O(w)$ from now on. The exact KKT condition at $w$ requires an index set $I(w)=\{\langle t,q \rangle~\big|~\text{dist}(p(t,w),q)=d_0\}$ on which:
\begin{equation}
\begin{aligned}
\label{eq:KKT}
0=&\nabla_w O(w)+\sum_{\langle t,q \rangle\in I}\lambda_{\langle t,q\rangle }\nabla_w \text{dist}(p(t,w),q)    \\
\ST&\lambda_{\langle t,q \rangle}\leq0.
\end{aligned}
\end{equation}
\prettyref{eq:KKT} is the exact KKT condition. But in practice, a solution satisfying \prettyref{eq:KKT} will never be reached because it requires a possibly infinite index set $I(w)$. Instead, we will show that our practical algorithm produces a solution that satisfies a perturbed KKT condition as defined below. We will show later that the perturbed KKT condition approaches \prettyref{eq:KKT} in the limit. 
\begin{definition}
A perturbed KKT condition, denoted as $\text{KKT}(\epsilon_1,\epsilon_2)$ where $\epsilon_1>0$ and $\epsilon_2>0$, could be defined on an inexact index set $\tilde{I}(w)=\{\langle t,q \rangle~\big|~d_0-\epsilon_1\leq \text{dist}(p(t,w),q)\leq d_0+\epsilon_1\}$ on which:
\begin{align*}
0=&\nabla_w O(w)+\sum_{\langle t,q \rangle\in \tilde{I}}\lambda_{\langle t,q \rangle}\nabla_w g_{\langle t,q \rangle}(w)    \\
\ST&\lambda_{\langle t,q \rangle}\leq0   
\\
 &
\|\nabla_w g_{\langle t,q \rangle}(w)-\nabla_w \text{dist}(p(t,w),q)\|\leq\epsilon_2,
\end{align*}
where $g_{\langle t,q \rangle}$ is some function of $w$ that satisfy the above conditions. In our setup, $g_{\langle t,q \rangle}(w)$ will be the distance offset by $d_0$ as shown in the next section.
\end{definition}
\begin{remark}
We have abused notation in the above KKT condition and used a summation over an (possibly) infinite set: $\langle t,q \rangle\in \tilde{I}$, which is well-defined due to the Carath\'{e}odory's theorem \cite{caratheodory1911variabilitatsbereich}. In other words, we could always choose a finite subset of $\tilde{I}$ and a corresponding set of Lagrangian multipliers such that the KKT condition still holds.
%https://math.stackexchange.com/questions/3759861/caratheodorys-theorem-for-vectors-in-a-cone
\end{remark}

\subsection{Solution Sequence}
We then show that \prettyref{alg:pipm} returns a solution that satisfies the perturbed KKT condition, assuming $\epsilon_g=0$. With each set of parameters $\lambda,x_0,\epsilon$, and from initial guess $w_0$, \prettyref{alg:pipm} will return a solution denoted as $w(\lambda,x_0,\epsilon,w_0)$. By introducing another parameter $\beta\in(0,1)$, we can construct an infinite sequence  $\{w_k~\big|~k\geq 1$ such that $w_k\triangleq w(\beta^k\lambda,\beta^kx_0,\beta^k\epsilon,w_{k-1})$. If we can solve P-IPM to local minimum, then the following function diminishes at $w_k$: 
\begin{equation}
\begin{aligned}
\label{eq:KKT_K}
0=&\nabla_w O(w_k)+\sum_{i\in S_k}\lambda_i^k\nabla g_i(w_k)  \\
\lambda_i^k=&\beta^k\lambda\nabla\text{clog}(g_i(w_k))\leq0,
\end{aligned}
\end{equation}
where $g_i(w)$ are functions of form $\text{dist}(\bullet,\bullet)-d_0$, and ${S}_k$ is the set of indices of non-zero log-barrier terms. From the definition of $\text{clog}$, it is obvious that $0\leq g_i(w_k)\leq \beta^kx_0$ for all $i\in S_k$. 

Based on \prettyref{Ass:A1}, it's obvious that $\{w_k~\big|~k\geq 1\}$ and $\{p(t,w_k)~\big|~t\in[0,T],k\geq 1\}$
are uniformly bounded. Therefore, the following lemma is obvious, which essentially shows that $w_k$ progressively approach $\text{KKT}$ with a diminishing error:
\begin{lemma}
Assuming \prettyref{Ass:A1}, $\text{dist}(x,y)$ is a Lipschitz-continuous function in $x$ with constant $L_1$, and $\nabla_x \text{dist}(x,y)$ is another Lipschitz-continuous function in $x$ with constant $L_2$; then
$w_k$ satisfies $\text{KKT}(L_1\beta^k\epsilon+\beta^k x_0,L_2\beta^k\epsilon)$.
\end{lemma}
\begin{IEEEproof}
We consider the special form of $g_i(w_k)=\text{dist}(\bullet,\bullet)-d_0$. Although $\bullet$ can be a point, a line segment, or a triangle, there must be a point on the sub-trajectory's convex hull $p_i^k$ and a point $q_i^k\in\mathcal{E}$ that realize $g_i$, i.e. $g_i(w_k)=\text{dist}(p_i^k,q_i^k)-d_0$ and $\nabla g_i(w_k)=\nabla_w \text{dist}(p_i^k,q_i^k)$. Further, due to our subdivision condition, there must be a point on the exact trajectory $p(t_i^k,w_k)$ such that $\|p_i^k-p(t_i^k,w_k)\|<\beta^k\epsilon$. In other words, each $g_i(w_k)$ is associated with three variables: a point on sub-trajectory's convex hull $p_i^k$, a point on the exact trajectory $p(t_i^k,w_k)$, and a point on the environment $q_i^k$. We have the following results:
\begin{align*}
&-L_1\beta^k\epsilon+d_0 \\
\leq&-\|\text{dist}(p(t_i^k,w_k),q_i^k)-\text{dist}(p_i^k,q_i^k)\|+\text{dist}(p_i^k,q_i^k)    \\
\leq&\text{dist}(p_i^k,q_i^k) \leq\text{dist}(p(t_i^k,w_k),q_i^k)  \\
\leq&\|\text{dist}(p(t_i^k,w_k),q_i^k)-\text{dist}(p_i^k,q_i^k)\|+\text{dist}(p_i^k,q_i^k)    \\
\leq&L_1\beta^k\epsilon+\beta^k x_0+d_0   \\
&\\&
\|\nabla g_i(w_k)-\nabla_w \text{dist}(p(t_i^k,w_k),q_i^k)\| \\
=&\|\nabla_w \text{dist}(p_i^k,q_i^k)-\nabla_w \text{dist}(p(t_i^k,w_k),q_i^k)\|
\leq L_2\beta^k\epsilon,
\end{align*}
so $w_k$ satisfies $\text{KKT}(L_1\beta^k\epsilon+\beta^k x_0,L_2\beta^k\epsilon)$ by choosing $\tilde{I}(w_k)=\{\langle t_i^k,q_i^k \rangle\}$, $g_{\langle t_i^k,q_i^k \rangle}=g_i$, and $\lambda_{\langle t_i^k,q_i^k \rangle}=\lambda_i^k$.
\end{IEEEproof}
Next, we consider a convergent subsequence
$\{w_{k(j)}~\big|~j \geq 1\}\to w_*$ (there must be at least one due to \prettyref{Ass:A1}). Although the exactness of KKT satisfaction at $w_{k(j)}$ can be made arbitrarily close to zero, we emphasize that $w_*$ might not satisfy KKT condition of \prettyref{eq:SIP}, of which a counterexample can be easily constructed at some $w_*$ without the linear independence constraint qualification (LICQ). Therefore, to establish first-order optimality at $w_*$, we need to design additional qualifications at local optima. Specifically we assume the boundedness of approximate Lagrangian multipliers:
\begin{assume}
\label{Ass:A2}
For a constant $L_3$ independent of $k$, we have $\|\nabla\text{clog}(g_i(w_k))\|\leq L_3$.
\end{assume}

\subsection{Subdivision Limit}
We now switch gears and show an intermediary result that, after subdivision, the number of log-barrier terms grows reciprocally with $\beta^k$. This result is due to the fact that the arc length of a B\'{e}zier curve is upper bounded by length of its control polygon according to \cite{gravesen1997adaptive}, which will be used to prove the KKT satisfaction at $w_*$.
\begin{lemma}
\label{Lem:terms}
Under \prettyref{Ass:A1}, there exists a constant $L_4$ independent of $k$, such that for any convergent subsequence
$\{w_{k(j)}~\big|~j\geq 1\}\to w_*$, the number of log-barrier terms involved in \prettyref{eq:KKT} is upper bounded by $L_4/\beta^{k(j)}$, i.e. $|S_{k(j)}|<L_4/\beta^{k(j)}$.
\end{lemma}
\begin{IEEEproof}
To show this result, we need to adopt a similar technique as \cite{gravesen1997adaptive}. We use the same notations as those in the main paper, i.e. the UAV trajectory is represented by $N$ B\'{e}zier curves each of order $M$ with parameter $s\in[0,1]$. For simplicity and without a loss of generality, we assume $N=1$. We introduce the arc-length of a B\'{e}zier curves as:
\begin{align*}
\Sigma(w)=\int_0^1\|\dot{A}(s)w\|ds.
\end{align*}
Notice that $\Sigma(w)$ is subdivision invariant. In other words, summing up $\Sigma(w)$ for each sub-curve after subdivision yields the arc-length of the entire curve. By the convergence of $w_{k(j)}$ and the continuity of $\Sigma(w)$, we can further choose a sufficiently large number $j_\Sigma$ such that $\Sigma(w_{k(j)})<2\Sigma(w_*)$ for all $j>j_\Sigma$. 

Next, we need to bound the arc-length of each sub-curve. By Cauchy-Schwarz inequality, we have:
\begin{equation}
\begin{aligned}
\label{eq:LBArc}
\Sigma(w)
=&\int_0^1\|\dot{A}(s)w\|ds\geq\sqrt{\int_0^1\|\dot{A}(s)w\|^2ds}   \\
=&\sqrt{w^T\left[\int_0^1\dot{A}(s)^T\dot{A}(s)ds\right]w}\triangleq\sqrt{w^THw}    \\
\triangleq&\sqrt{w^TD^T\bar{H}Dw}\geq\sigma_{\min}(\sqrt{\bar{H}})\|Dw\|.
\end{aligned}
\end{equation}
Here $D$ is the stencil that extracts the edges of control polygon, defined as:
\begin{align*}
D\triangleq
\left(\setlength{\arraycolsep}{1pt}
\begin{array}{ccccc}
I & -I & & &  \\
& I & -I & &  \\
&& \ddots & \ddots &  \\
& & & I & -I  \\
\end{array}\right).
\end{align*}
Since the derivative of a B\'{e}zier curve is another B\'{e}zier curve with control points being $Dw$, we must have $H=D^T\bar{H}D$ for some positive semi-definite matrix $\bar{H}$. We further emphasize that $\sigma_{\min}(\sqrt{\bar{H}})>0$, i.e. $\bar{H}$ is strictly positive definite. This is because the derivative of a B\'{e}zier curve of order $M$ is another B\'{e}zier curve of order $M-1$, with control points being $Dw$. In addition, a B\'{e}zier curve is uniquely defined by its control polygons \cite{berry1997uniqueness}, so that if $w^TD^T\bar{H}Dw=0$, then $Dw=0$, which further implies that $\bar{H}$ has a full rank.

We then bound the term $\|Dw\|$, which can be done using our subdivision rule. We first notice that $\|Dw\|$ is the total length of control polygon and we know that the total length of control polygon must be larger than the diameter of a control polygon $\Delta(A,w)$. (This is because the diameter of a polytope is realized on two vertices, the distance between which is smaller than the path connecting all vertices.) We consider a subdivision that is done on $A(s)w$ to derive two sub-curves with control points: $D_{1,2}w$, respectively. If $\|DD_1w\|\leq \beta^{k(j)}\epsilon\min(\sigma_{\min}(D_1),\sigma_{\min}(D_2))$, then
\begin{align*}
&\Delta(A,w)\leq\|Dw\|\leq\|DD_1w\|/\sigma_{\min}(D_1)  \\
\leq&\beta^{k(j)}\epsilon\frac
{\min(\sigma_{min}(D_1),\sigma_{\min}(D_2))}{\sigma_{\min}(D_1)}
\leq\beta^{k(j)}\epsilon,
\end{align*}
and a similar result holds for $\|DD_2w\|$. However, it is impossible for $\Delta(A,w)\leq\beta^{k(j)}\epsilon$ because our subdivision rule will prevent further subdivision. As a result, we must have that every curve that is subdivided to the limit satisfies the following condition:
\begin{align*}
\|Dw\|\geq\beta^{k(j)}\epsilon\min(\sigma_{\min}(D_1),\sigma_{\min}(D_2)).
\end{align*}
Combining this result with \prettyref{eq:LBArc}, we have the following lower bound of the arc-length of each subdivided B\'ezier curve piece:
\begin{align}
\label{eq:LBArcEps}
\beta^{k(j)}\epsilon\sigma_{\min}(\sqrt{\bar{H}})
\min(\sigma_{\min}(D_1),\sigma_{\min}(D_2)).
\end{align}
Finally, we can define $L_4$ as follows:
\small
\begin{align*}
L_4\triangleq&\left\lceil\frac{2\Sigma(w_*)}
{\epsilon\sigma_{\min}(\sqrt{\bar{H}})\min(\sigma_{\min}(D_1),\sigma_{\min}(D_2))}\right\rceil\\
&\left[|\mathcal{L}|\frac{(M+1)M}{2}+
|\mathcal{P}|\frac{(M+1)M(M-1)}{6}+
|\mathcal{T}|(M+1)\right],
\end{align*}
\normalsize
and prove by contradiction. If we can find infinitely many $j$ such that $|S_{k(j)}|\geq L_4/\beta^{k(j)}$. Note that the distance between a convex hull and the environment boils down to finitely many distance between geometric primitive pairs, where the number of terms are upper bounded by:
\begin{align*}
|\mathcal{L}|\frac{(M+1)M}{2}+
|\mathcal{P}|\frac{(M+1)M(M-1)}{6}+
|\mathcal{T}|(M+1).
\end{align*}
By the Pigeonhole principle, the number of subdivided B\'ezier curve pieces is at least:
\begin{align}
\label{eq:L4}
\left\lfloor
\frac{1}{\beta^{k(j)}}\left\lceil\frac{2\Sigma(w_*)}
{\epsilon\sigma_{\min}(\sqrt{\bar{H}})\min(\sigma_{\min}(D_1),\sigma_{\min}(D_2))}\right\rceil
\right\rfloor.
\end{align}
By \prettyref{eq:LBArcEps} and \prettyref{eq:L4}, the total arc-length satisfies $\Sigma(w_{k(j)})\geq2\Sigma(w_*)$ for infinitely many $j$, which contradicts the convergence of $\Sigma(w_{k(j)})$ to $\Sigma(w_*)$.
\end{IEEEproof}

\subsection{KKT Condition at $w_*$}
We can finally prove by contradiction that KKT condition (\prettyref{eq:KKT}) holds at $w_*$. It is obvious that $w_*$ is a feasible point of \prettyref{eq:SIP}. Suppose otherwise that KKT condition does not hold at $w_*$, then there must be a unit direction $v$ such that:%\xifeng{still don't get it}:
\small
\begin{equation}
\begin{aligned}
\label{eq:dir}
&v^T\nabla_w O(w_*)=\epsilon_3<0  \\
&v^T\nabla_w\text{dist}(p(t,w_*),q)\geq0\quad\forall \langle t,q \rangle\in I(w_*).
\end{aligned}
\end{equation}
\normalsize
\begin{remark}
\label{rem:KKTFailure}
To show that \prettyref{eq:dir} holds whenever the KKT condition (\prettyref{eq:KKT}) fails, let us consider the convex cone $\mathbb{C}=\{\sum_{\langle t,q \rangle\in I}\lambda_{\langle t,q\rangle }\nabla_w \text{dist}(p(t,w),q)|\lambda_{\langle t,q\rangle }\leq0\}$. The failure of \prettyref{eq:KKT} implies that $-\nabla_wO$ does not belong to $\mathbb{C}$. If $\mathbb{C}$ is closed, we can then project $-\nabla_wO$ back to the cone along some normal vector $v$, which is the vector we are looking for in \prettyref{eq:dir} (see e.g., \cite{rockafellar2015convex}).
\end{remark}
\begin{remark}
Informally, we take three steps to show that $\mathbb{C}$ in \prettyref{rem:KKTFailure} is closed. First, we define another set $\bar{\mathbb{C}}=\{\nabla_w\text{dist}(p(t,w),q)|\langle t,q \rangle\in I\}$, which is obviously bounded. We argue that $\bar{\mathbb{C}}$ is also closed. This is because a B\'{e}zier curve can only touch a triangle mesh in finitely many closed segments, $\nabla_w\text{dist}$ is a continuous function, and a finite union of closed sets is again closed. We conclude that $\bar{\mathbb{C}}$ is compact. Second, we have $\mathbb{C}=\text{Cone}(\text{Conv}(-\bar{\mathbb{C}}))$ and the convex hull of a compact set is compact in $\mathbb{R}^n$ so $\text{Conv}(-\bar{\mathbb{C}})$ is compact and convex. Finally, the cone of a compact convex set is closed so $\mathbb{C}$ is closed.
%https://math.stackexchange.com/questions/2811835/in-rn-convex-cone-of-any-compact-set-is-closed-cone-of-any-compact-set-is-c
%https://math.stackexchange.com/questions/3145466/convex-hull-of-a-compact-in-rn-is-compact
\end{remark}

We prove that $v$ must be a descend direction at $w_{k(j)}$ for sufficiently large $j$, contradicting the fact that $w_{k(j)}$ is a local minimum of P-IPM. Note that for any point $p(t_i^{k(j)},w_{k(j)})$, it is not possible for $\text{dist}(p(t_i^{k(j)},w_{k(j)}),q_i^{k(j)})=d_0$ as the log-barrier will be infinite otherwise, i.e. we cannot reach any point in the index set of $w_*$, but we can bound their distance using the following lemma:
\begin{lemma}
\label{Lem:INDEX_APPROACH}
Under \prettyref{Ass:A1}, given any $\epsilon_4>0$, there is a large enough $j(\epsilon_4)$, such that for all $j\geq j(\epsilon_4)$ and $\text{clog}(g_i(w_{k(j)}))>0$, we can find some $\langle t,q \rangle\in I(w_*)$ satisfying:
\begin{align*}
\|p(t_i^{k(j)},w_{k(j)})-p(t,w_*)\|\leq\epsilon_4.
\end{align*}
\end{lemma}
\begin{IEEEproof}
We proof by contradiction. Suppose otherwise, there exists some $\epsilon_4>0$ such that for any $j_0$, we can always find some $j>j_0$ and $\text{clog}(g_{i(j)}(w_{k(j)}))>0$ where $\|p(t_{i(j)}^{k(j)},w_{k(j)})-p(t,w_*)\|>\epsilon_4$ for all $\langle t,q \rangle\in I(w_*)$. Then we can construct an infinite sub-sequence of form $\{p(t_{i(j)}^{k(j)},w_{k(j)})\}$ in which every point is at least $\epsilon_4$ away from the index set $I(w_*)$. Due to \prettyref{Ass:A1}, there will be a convergent subsequence of $\{p(t_{i(j)}^{k(j)},w_{k(j)})\}$, the limit of which denoted as $p(t_*,w_*)$. We have $\text{dist}(p(t_*,w_*),\mathcal{E})=0$ so $p(t_*,w_*)$ belongs to the index set. Further, $p(t_*,w_*)$ is at least $\epsilon_4$ away from the index set, a contradiction. 
\end{IEEEproof} 
We can now present our main result, which involves bounding each term in \prettyref{eq:KKT_K}: 
\begin{proposition}
\label{Prop:INDEX_APPROACH}
Under \prettyref{Ass:A1} and \prettyref{Ass:A2}, for any convergent subsequence $\{w_{k(j)}~\big|~j \geq 1\}\to w_*$, a unit vector satisfying \prettyref{eq:dir} is a descend direction at $w_{k(j)}$ for sufficiently large $j$.
\end{proposition}
\begin{IEEEproof}
We have bounded $w$ under \prettyref{Ass:A1} and $t\in[0,T]$ is also bounded. As a result, a point on the curve, namely $p(t,w)$, is a continuous function on the bounded domain, which is also Lipschitz continuous. For the same reason, $A(t)$ is Lipschitz continuous. We denote the Lipschitz constant for these two functions as $L_5$. We start our proof by bounding the constraint gradient $\nabla g_i$. We divide the directional derivative of $\nabla g_i$ along $v$ into three terms, each equipped with a diminishing bound:
\small
\begin{align*}
 &v^T\nabla g_i(w_{k(j)})    \\
=&v^T(\nabla g_i(w_{k(j)})-\nabla_w \text{dist}(p(t_i^{k(j)},w_{k(j)}),q_i^{k(j)}))+   \\
&\text{(first term)}    \\
 &v^T(\nabla_w \text{dist}(p(t_i^{k(j)},w_{k(j)}),q_i^{k(j)})-\nabla_w \text{dist}(p(t,w_*),q))+   \\
 &\text{(second term)}   \\
 &v^T\nabla_w \text{dist}(p(t,w_*),q)\quad\text{(third term)} \\
\geq&-L_2\beta^{k(j)}\epsilon-L_2\|p(t_i^{k(j)},w_{{k(j)}})-p(t,w_*)\|    \\
\geq&-L_2\beta^{k(j)}\epsilon-L_2\epsilon_4,
\end{align*}
\normalsize
For the first term above, we use our subdivision rule and continuity of $\nabla_w \text{dist}$. For the second term, we use \prettyref{Lem:INDEX_APPROACH} by choosing $j>j(\epsilon_4)$. The third term is bigger than zero due to $v$ violating the KKT condition (\prettyref{eq:dir}). Next, we can invoke \prettyref{Lem:terms} and show that:
\begin{align*}
v^T\sum_{i\in S_{k(j)}}\lambda_i^{k(j)}\nabla g_i(w_{k(j)})\leq L_3L_4\lambda(L_2\beta^{k(j)}\epsilon+L_2\epsilon_4).
\end{align*}
Finally, our algorithm only consider twice-differentiable objective function $O$. Under assumption \prettyref{Ass:A1}, $\nabla_wO$ is Lipschitz continuous in the bounded domain of $w$ with constant $L_6$. As a result, we have:
\small
\begin{align*}
v^T\nabla_wO(w_{k(j)})=&v^T(\nabla_wO(w_{k(j)})-\nabla_wO(w_*))+v^T\nabla_wO(w_*)   \\
\leq&\epsilon_3+L_6\|w_{k(j)}-w_*\|.
\end{align*}
\normalsize
Combining all the results above, we conclude that:
\begin{align*}
&v^T\nabla_w\left[O(w_{k(j)}+\lambda B(w_{k(j)},\mathcal{E})\right] \\
\leq&\epsilon_3+L_6\|w_{k(j)}-w_*\|+
L_3L_4\lambda(L_2\beta^{k(j)}\epsilon+L_2\epsilon_4),
\end{align*}
which can be made arbitrarily close to $\epsilon_3$ by choosing large enough $j$, contradicting the fact that $w_{k(j)}$ is a local minimum of P-IPM.
\end{IEEEproof}
We can now claim our main result:
\begin{corollary}
Any accumulation point of the sequence $\{w_k\}$ satisfies the KKT condition of SIP (\prettyref{eq:KKT}) under \prettyref{Ass:A1} and \prettyref{Ass:A2}.
\end{corollary}

%% file: 08-appendixB.tex
\section{Finite Termination of Exact P-IPM\label{sec:pipm-proof2}}
In the above convergence analysis of exact P-IPM to SIP, we assume that $\epsilon_g=0$. However, a practical algorithm needs to have finite termination. In this section, we show that as long as $\epsilon_g>0$, the exact P-IPM \prettyref{alg:pipm} always converges in a finite number of iterations. Finite termination happens under two conditions:
\begin{itemize}
\item The while loop in line-search \prettyref{alg:search} is finite.
\item The outer loop in \prettyref{alg:pipm} is finite.
\end{itemize}
Although the finite termination of Newton's method has been shown in \cite{bertsekas1997nonlinear}, we have used a special safeguard in our line search (\prettyref{ln:safeguard} in \prettyref{alg:search}), which is a new setting whose convergence needs to be shown. 

\subsection{Finite Termination of Line Search}
\begin{lemma}
\label{lem:searchTerminate}
\prettyref{alg:search} terminates after finitely many iterations, if started from a $w$ with finite objective function value.
\end{lemma}
\begin{IEEEproof}
Consider any point $p\in\text{Hull}(\mathcal{P}'(w))$, which is a convex combination of its $N+1$ vertices. If we update $w$ to $w'=w+\alpha d$ along some descendent direction $d$, then $p$ will be moved to $p'$ with $\text{dist}(p,p')\leq\alpha L_1 L_5\|d\|$. If we choose:
\begin{align}
\label{eq:upper_bound}
\alpha\leq\frac{\text{dist}(\text{Hull}(\mathcal{P}'(w)),\mathcal{E})-d_0}{2L_1L_5\|d\|},
\end{align}
then we have:
\begin{align*}
&\text{dist}(l_{pp'}',\mathcal{E})\\
\geq&\text{dist}(p,\mathcal{E})-\text{dist}(p,p')\\
\geq&\frac{\text{dist}(\text{Hull}(\mathcal{P}'(w)),\mathcal{E})+d_0}{2}>d_0,
\end{align*}
where $l_{pp'}'$ is a line segment connecting $p$ and $p'$. Since $p$ is an arbitrary point in the convex hull, we thus have:
\begin{align*}
\text{dist}(\text{Hull}(\mathcal{P}'(w),\mathcal{P}'(w')),\mathcal{E})>d_0,
\end{align*}
which implies that the safeguard will pass as long as \prettyref{eq:upper_bound} holds. Due to the finite objective function value, the righthand side of \prettyref{eq:upper_bound} is positive. Then by the differentiability of the objective function, a positive $\alpha$ satisfying Wolfe's first condition can be found.
\end{IEEEproof}

\subsection{Finite Termination of Outer Loop\label{subsec:pipm-outloop}}
We have the following corollary of this assumption based on \prettyref{Ass:A1}:
\begin{corollary}
\label{Cor:A}
Suppose \prettyref{Ass:A1} holds ($w$ is bounded). For each $w$ generated by \prettyref{alg:pipm}, we have:
\begin{align*}
\text{dist}(\text{Hull}(\mathcal{P}'(w)),\mathcal{E})\geq d_0+L_7,
\end{align*}
with some $L_7>0$ independent of iteration number.
\end{corollary}
\begin{IEEEproof}
If no such $L_7$ can be found, then $\text{dist}(\text{Hull}(\mathcal{P}'(w)),\mathcal{E})-d_0<L_7$ for any sufficiently small $L_7$. If we choose a sequence of $L_7\to0$, we have the objective function $O(w,T)+\lambda B(w,\mathcal{E})\geq\lambda\text{clog}(L_7)\to \infty$. This contradicts the monotonic decrease of the objective function.
\end{IEEEproof}
\begin{corollary}
\label{Cor:B}
Suppose \prettyref{Ass:A1} holds. For each $w$ generated by \prettyref{alg:pipm}, we have:
\begin{align*}
\left\|\FPP{O(w,T)}{w}+\lambda\FPP{B(w,\mathcal{E})}{w}\right\|<L_8,
\end{align*}
with some $L_8>0$ independent of iteration number.
\end{corollary}
\begin{IEEEproof}
Since $w$ is bounded and $O$ is twice-differentiable, its gradient is bounded. By a similar argument as \prettyref{Lem:terms}, we have that the number of subdivided B\'ezier curve pieces is at most:
\begin{align*}
\left\lceil
\frac{\|Dw\|\sigma_{\max}(\sqrt{\bar{H}})}
{\epsilon\sigma_{\min}(\sqrt{\bar{H}})\min(\sigma_{\min}(D_1),\sigma_{\min}(D_2))}
\right\rceil
\end{align*}
Combining these two results, we have:
\small
\begin{align*}
&\left\|\FPP{O(w,T)}{w}+\lambda\FPP{B(w,\mathcal{E})}{w}\right\|   \\
\leq&\left\|\FPP{O(w,T)}{w}\right\|
+\lambda 
\left\lceil
\frac{\|Dw\|\sigma_{\max}(\sqrt{\bar{H}})}
{\epsilon\sigma_{\min}(\sqrt{\bar{H}})\min(\sigma_{\min}(D_1),\sigma_{\min}(D_2))}
\right\rceil\\
&\left[|\mathcal{L}|\frac{(M+1)M}{2}+
|\mathcal{P}|\frac{(M+1)M(M-1)}{6}+
|\mathcal{T}|(M+1)\right]\\
&\left|\FPP{\text{clog}(d)}{d}\Bigg|_{d=d_0+L_7}\right|
\left\|\FPP{\text{dist}(p(t,w),\bullet)}{w}\right\|,
\end{align*}
\normalsize
where $\bullet$ means any geometric primitive and we have used \prettyref{Cor:A} to bound $\|Dw\|$. Further, it can be shown that the gradient norm $\FPPR{\text{clog}(d)}{d}$ is monotonically decreasing in $d$, so it can be lower-bounded to the gradient norm at $d_0+L_7$ due to \prettyref{Cor:A}. $\FPPR{O(w,T)}{w}$ is bounded by $L_6$ due to \prettyref{Ass:A1}. $\FPPR{\text{dist}(p(t,w))}{w}$ is bounded because:
\begin{align*}
\left\|\FPP{\text{dist}(p(t,w))}{w}\right\|\leq
\left\|\FPP{\text{dist}(p(t,w))}{p(t,w)}\right\|
\left\|\FPP{p(t,w)}{w}\right\|\leq L_2L_5.
\end{align*}
\end{IEEEproof}

\begin{IEEEproof}[Proof of \prettyref{Prop:ExactTerminate}]
Similar to the proof of \prettyref{lem:searchTerminate}, we consider every point $p\in\text{Hull}(\mathcal{P}'(w))$, which is a convex combination of its $N+1$ vertices. If we update $w$ to $w'=w+\alpha d$, then $p$ will be moved to $p'$ with $\text{dist}(p,p')\leq\alpha L_1 L_5 L_8$, where we have used \prettyref{Cor:B} to bound $d$. By \prettyref{Cor:A} we have $\text{dist}(l_{pp'}',\mathcal{E})>d_0+L_7$, so triangle inequality implies that:
\begin{align*}
\text{dist}(l_{pp'}',\mathcal{E})\geq
\text{dist}(p,\mathcal{E})-\text{dist}(p,p')\geq 
d_0+L_7-\alpha L_1 L_5 L_8. 
\end{align*}
Therefore, our line search will always pass the safeguard test, if we choose $\alpha\leq L_7/(L_1L_5L_8)$. The finite termination follows by choosing a stepsize sequence with each term smaller than $L_7/(L_1L_5L_8)$ and \cite[Proposition~1.2.4]{bertsekas1997nonlinear}.
\end{IEEEproof}
Note that we have only analyzed gradient descend method (i.e. $\text{SPD}(\nabla_{w,T}^2 O)=I$), the case with Newton's method is almost identical, assuming the Hessian matrix has bounded singular values.

%% file: 09-appendixC.tex
\section{Finite Termination of Inexact P-IPM\label{sec:inpipm}}
The convergence analysis for inexact P-IPM is more obscure. In this setting, the safeguard algorithm measures the distance based on convex hull, while the objective function simplifies the convex hull to a line segment. This inconsistent geometric representation might invalidate finite termination in the line-search \prettyref{alg:search}. In the following sections, we prove that the inexact P-IPM algorithm has finite termination. The core to our proof is the derivation of the error between $\tilde{B}$ and $\hat{B}$ can be bounded.
\begin{remark}
Our definition of $\tilde{B}$ and $\hat{B}$ is not differentiable for general environment mesh $\mathcal{E}$. A differentiable version can be derived by breaking up $\mathcal{E}$ into edges and triangles. For simplicity of proof, we assume that $\mathcal{E}$ is a single edge throughout this section. The proof can be applied to general environment meshes by adding separate terms to both $\tilde{B}$ and $\hat{B}$ for each edge and triangle.
\end{remark}

\subsection{Well-Defined Integral of Log-Barrier}
We have used a different version of $\text{clog}$ from its original definition in \cite{10.1145/3386569.3392425}, for which the reason is to ensure the well-definedness of $\hat{B}$. A well-defined log-barrier function should take infinite value when any part of the B\'ezier curve intersects $\mathcal{E}$. We present a counter example in \prettyref{fig:counterExample} to show that the original $\text{log}$ function in \cite{10.1145/3386569.3392425} does not pertain this property. Indeed, the value of $\hat{B}$ in \prettyref{fig:counterExample} can be evaluated analytically as follows:
\begin{align*}
&\int_0^1-(|s-0.5|-0.5)^2\text{log}\frac{|s-0.5|}{0.5}ds=\frac{11}{72}<\infty,
\end{align*}
where we choose $x_0=0.5$. For our modified $\text{clog}$ function, we prove well-definedness in the following lemma:
\begin{lemma}
If $\text{dist}(A(s)w,\mathcal{E})=d_0$ for some $s\in(0,1)$, then $\hat{B}=\infty$.
\end{lemma}
\begin{IEEEproof}
Consider the closed interval $[s-\epsilon_5,s+\epsilon_5]\subset(0,1)$ for some sufficiently small $\epsilon_5$. The value of $\text{dist}$ is upper bounded by:
\begin{align*}
0\leq\text{dist}(A(s')w,\mathcal{E})\leq L_1L_5\epsilon_5+d_0,
\end{align*}
 where $s' \in [s-\epsilon_5,s+\epsilon_5]$.
The sub-integral of $\hat{B}$ in range $[s-\epsilon_5,s+\epsilon_5]$ is thus lower bounded by:
\begin{align*}
\hat{B}(w,\mathcal{E})\geq&\int_{s-\epsilon_5}^{s+\epsilon_5}\text{clog}(\text{dist}(A(s)w,\mathcal{E})-d_0)ds\\
\geq&2L_1L_5\epsilon_5\text{clog}(L_1L_5\epsilon_5)\\
=&-2(L_1L_5\epsilon_5-x_0)^2\text{log}(\frac{L_1L_5\epsilon_5}{x_0}),
\end{align*}
the righthand side of which tends to $\infty$ as $\epsilon_5\to0$.
\end{IEEEproof}
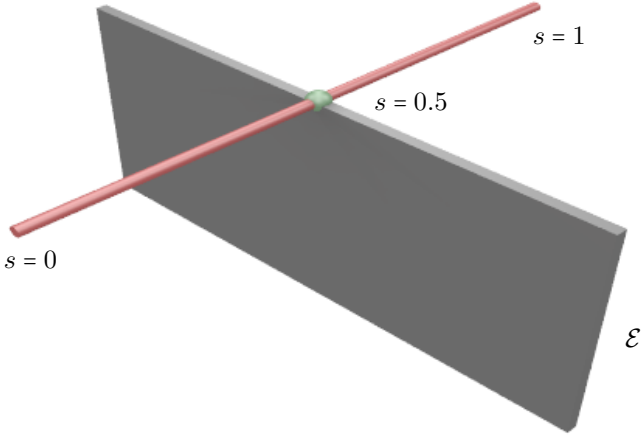
\begin{figure}[ht]
\centering
\includegraphics[width=.95\columnwidth]{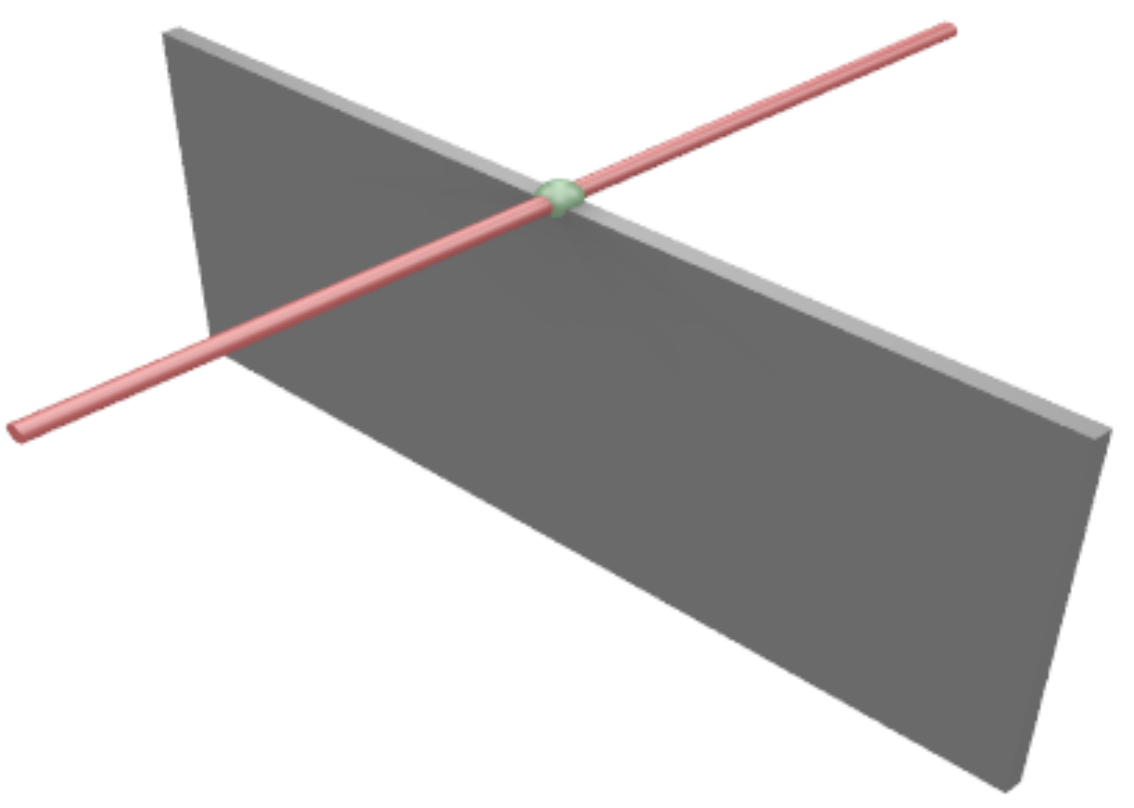}
\put(-5,35){$\mathcal{E}$}
\put(-240,65){$s=0$}
\put(-100,125){$s=0.5$}
\put(-40,150){$s=1$}
\caption{\label{fig:counterExample} In this counter example, the red line is the UAV's B\'ezier curve and the gray horizontal wall is the environment, touching at exactly one green point. However, $\hat{B}$ takes a finite value if $\text{clog}$ from \cite{10.1145/3386569.3392425} is used.}
\end{figure}

\subsection{Adaptive Subdivision}
A potential issue in using $\hat{B}$ for our analysis is that $\tilde{B}$ approaches $\hat{B}$ only under uniform subdivisions (i.e. subdividing every curve piece from $s_i$ to $s_{i+1}$). However, we use adaptive subdivisions. To bridge this gap, we need the following definition for further analysis:
\begin{align*}
&\hat{B}_{\epsilon_6}(w,\mathcal{E})\triangleq\sum_{i=1}^{S-1}\\
&\begin{cases}
\int_{s_i}^{s_{i+1}}\text{clog}(\text{dist}(A(s)w,\mathcal{E})-d_0)ds&s_{i+1}-s_i<\epsilon_6 \\
(s_{i+1}-s_i)\text{clog}(\text{dist}(m_{i,i+1}'(w),\mathcal{E})-d_0)&s_{i+1}-s_i\geq\epsilon_6.
\end{cases}
\end{align*}
Intuitively, $\hat{B}_{\epsilon_6}(w,\mathcal{E})$ combines $\tilde{B}$ and $\hat{B}$ based on the length of each subdivided curve piece. A pivotal property of $\hat{B}$ is that it is invariant to subdivision. Similarly, we can show that $\hat{B}_{\epsilon_6}$ is invariant to subdivision after sufficiently large number of calls to \prettyref{alg:subdUnsafe}:
\begin{lemma}
\label{Lem:BHatInvariance}
For any fixed $\epsilon_6$, after sufficiently but finitely many calls to \prettyref{alg:subdUnsafe}, $\hat{B}_{\epsilon_6}$ is invariant to subdivision.
\end{lemma}
\begin{IEEEproof}
It takes finitely many calls to \prettyref{alg:subdUnsafe} to bring every $s_{i+1}-s_i<\epsilon_6$.
\end{IEEEproof}
We can now bound the error between $\tilde{B}$ and $\hat{B}_{\epsilon_6}$ by controlling $\epsilon_6$ as follows:
\begin{lemma}
\label{Lem:ValueErr}
Taking \prettyref{Ass:A1}, if $w$ is generated by some iteration of \prettyref{alg:pipmInexact}, then we have the following bound:
\small
\begin{align*}
|\tilde{B}-\hat{B}_{\epsilon_6}|\leq\epsilon_6
\left[L_9\frac{\text{log}(\psi(\epsilon_6))}{\psi(\epsilon_6)^2}+L_{10}\right]L_1L_5
=\mathcal{O}(\epsilon_6^{1-2\eta}\text{log}(\epsilon_6)),
\end{align*}
\normalsize
for some negative constant $L_9$ and positive constant $L_{10}$. The bound is diminishing as $\epsilon_6\to0$ if $\eta<1/2$. 
\end{lemma}
\begin{IEEEproof}
If $w$ is generated by some iteration of \prettyref{alg:pipmInexact}, then every B\'ezier curve piece must have passed the safeguard check inside the line search. For a curve piece satisfying $s_{i+1}-s_i\geq\epsilon_6$, $\tilde{B}$ and $\hat{B}_{\epsilon_6}$ are the same on that segment. For a curve piece satisfying $s_{i+1}-s_i<\epsilon_6$, instead, we have from the safeguard condition that:
\begin{align*}
\text{dist}(A(s)w,\mathcal{E})-d_0\geq\psi(s_{i+1}-s_i)\quad\forall s\in[s_i,s_{i+1}].
\end{align*}
For any two parameters $s_1,s_2\in[s_i,s_{i+1}]$, we have from the mean value theorem that:
\small
\begin{align}
\label{eq:ErrBoundInte}
&|\text{clog}(\text{dist}(A(s_1)w,\mathcal{E})-d_0)-\text{clog}(\text{dist}(A(s_2)w,\mathcal{E})-d_0)|\nonumber\\
=&\left|\int_{\text{dist}(A(s_1)w,\mathcal{E})-d_0}^{\text{dist}(A(s_2)w,\mathcal{E})-d_0}
\FPP{\text{clog}(x)}{x}dx\right|\nonumber\\
\leq&\left|\FPP{\text{clog}(d)}{d}\Bigg|_{\psi(s_{i+1}-s_i)}\right|
|\text{dist}(A(s_1)w,\mathcal{E})-\text{dist}(A(s_2)w,\mathcal{E})|\nonumber\\
\leq&\left|\FPP{\text{clog}(d)}{d}\Bigg|_{\psi(s_{i+1}-s_i)}\right||A(s_1)w-A(s_2)w|L_1\nonumber\\
\leq&|s_1-s_2|\left|\FPP{\text{clog}(d)}{d}\Bigg|_{\psi(s_{i+1}-s_i)}\right|L_1L_5.
\end{align}
\normalsize
It can be verified that as $d\to0$, the dominating term of $\FPPR{\text{clog}(d)}{d}$ is $\text{log}(d)/d^2$. Therefore, we can find negative constant $L_9$ and positive constant $L_{10}$ such that: 
\begin{align*}
\left|\FPP{\text{clog}(d)}{d}\right|\leq L_9\frac{\text{log}(d)}{d^2}+L_{10}\quad\forall d>0,
\end{align*}
which can be plugged into \prettyref{eq:ErrBoundInte} to derive:
\begin{align*}
&|\text{clog}(\text{dist}(A(s_1)w,\mathcal{E})-d_0)-\text{clog}(\text{dist}(A(s_2)w,\mathcal{E})-d_0)|\\
\leq&|s_1-s_2|\left[L_9\frac{\text{log}(\psi(s_{i+1}-s_i))}{\psi(s_{i+1}-s_i)^2}+L_{10}\right]L_1L_5.
\end{align*}
Therefore, the difference between $\tilde{B}$ and $\hat{B}_{\epsilon_6}$ can be bounded on the curve piece as:
\begin{align*}
&\bigg|\int_{s_i}^{s_{i+1}}\text{clog}(\text{dist}(A(s)w,\mathcal{E})-d_0)ds\\
&-(s_{i+1}-s_i)\text{clog}(\text{dist}(m_{i,i+1}'(w),\mathcal{E})-d_0)\bigg|\\
\leq&(s_{i+1}-s_i)^2
\left[L_9\frac{\text{log}(\psi(s_{i+1}-s_i))}{\psi(s_{i+1}-s_i)^2}+L_{10}\right]L_1L_5\\
\leq&(s_{i+1}-s_i)\epsilon_6
\left[L_9\frac{\text{log}(\psi(\epsilon_6))}{\psi(\epsilon_6)^2}+L_{10}\right]L_1L_5.
\end{align*}
The difference between $\tilde{B}$ and $\hat{B}_{\epsilon_6}$ is maximized when all the curve pieces satisfy $s_{i+1}-s_i<\epsilon_6$. Combined with the fact that $\sum_{i=1}^{S-1}(s_{i+1}-s_i)=1$, we derive the bound to be proved. Further, if $\eta<1/2$, then the leading term of $|\tilde{B}-\hat{B}_{\epsilon_6}|$ is  $\epsilon_6^{1-2\eta}\text{log}(\epsilon_6)$, which is diminishing as $\epsilon_6\to0$.
\end{IEEEproof}
Following a similar reasoning, we can also bound the different in gradient:
\begin{lemma}
\label{Lem:GradErr}
Taking \prettyref{Ass:A1}, if $w$ is generated by some iteration of inexact P-IPM, then we have the following bound:
\begin{align*}
&\|\nabla_w\tilde{B}-\nabla_w\hat{B}_{\epsilon_6}\|\\
\leq&
\epsilon_6\left[L_9\frac{\text{log}(\psi(\epsilon_6))}{\psi(\epsilon_6)^2}+L_{10}\right]
(L_2L_5\fmax{s\in[0,1]}\|A(s)\|+L_5)+\\
&\epsilon_6\left[L_{11}\frac{\text{log}(\psi(\epsilon_6))}{\psi(\epsilon_6)^3}+L_{12}\right]
(L_2L_5\fmax{s\in[0,1]}\|A(s)\|)\\
=&\mathcal{O}(\epsilon_6^{1-3\eta}\text{log}(\epsilon_6)),
\end{align*}
for some negative constant $L_{11}$ and positive constant $L_{12}$. The bound is diminishing as $\epsilon_6\to0$ if $\eta<1/3$.
\end{lemma}
\begin{IEEEproof}
For a curve piece satisfying $s_{i+1}-s_i\leq\epsilon_6$ and two parameters $s_1,s_2\in[s_i,s_{i+1}]$, we have from the triangle inequality that:
\small
\begin{align*}
&\|\nabla_w\text{clog}(\text{dist}(A(s_1)w,\mathcal{E})-d_0)-
\nabla_w\text{clog}(\text{dist}(A(s_2)w,\mathcal{E})-d_0)\|\\
=&\left\|\FPP{\text{clog}(d)}{d_1}\FPP{\text{dist}(p,\mathcal{E})}{p_1}A(s_1)
-\FPP{\text{clog}(d)}{d_2}\FPP{\text{dist}(p,\mathcal{E})}{p_2}A(s_2)\right\|\\
=&\|\FPP{\text{clog}(d)}{d_1}\FPP{\text{dist}(p,\mathcal{E})}{p_1}A(s_1)-
\FPP{\text{clog}(d)}{d_1}\FPP{\text{dist}(p,\mathcal{E})}{p_2}A(s_2)+\\
&\FPP{\text{clog}(d)}{d_1}\FPP{\text{dist}(p,\mathcal{E})}{p_2}A(s_2)-
\FPP{\text{clog}(d)}{d_2}\FPP{\text{dist}(p,\mathcal{E})}{p_2}A(s_2)\|\\
\leq&\left|\FPP{\text{clog}(d)}{d_1}\right|
\left\|\FPP{\text{dist}(p,\mathcal{E})}{p_1}A(s_1)-\FPP{\text{dist}(p,\mathcal{E})}{p_2}A(s_2)\right\|+\\
&\left|\FPP{\text{clog}(d)}{d_1}-\FPP{\text{clog}(d)}{d_2}\right|
\left\|\FPP{\text{dist}(p,\mathcal{E})}{p_2}A(s_2)\right\|,
\end{align*}
\normalsize
where $d_1=\text{dist}(p_1,\mathcal{E})$, $p_1=p(s_1,w)=A(s_1)w$, and we have the same for $d_2,p_2$. There are two terms in the above equation to be bounded. For the first term, we have:
\small
\begin{align*}
&\left|\FPP{\text{clog}(d)}{d_1}\right|
\left\|\FPP{\text{dist}(p,\mathcal{E})}{p_1}A(s_1)-\FPP{\text{dist}(p,\mathcal{E})}{p_2}A(s_2)\right\|\\
\leq&\left|\FPP{\text{clog}(d)}{d_1}\right|
\left\|\FPP{\text{dist}(p,\mathcal{E})}{p_1}-\FPP{\text{dist}(p,\mathcal{E})}{p_2}\right\|\|A(s_1)\|+\\
&\left|\FPP{\text{clog}(d)}{d_1}\right|
\left\|\FPP{\text{dist}(p,\mathcal{E})}{p_2}\right\|\left\|A(s_1)-A(s_2)\right\|\\
\leq&\left|\FPP{\text{clog}(d)}{d_1}\bigg|_{\psi(s_{i+1}-s_i)}\right|L_2L_5(s_{i+1}-s_i)\fmax{s\in[0,1]}\|A(s)\|+\\
&\left|\FPP{\text{clog}(d)}{d_1}\bigg|_{\psi(s_{i+1}-s_i)}\right|L_5(s_{i+1}-s_i)\\
\leq&(s_{i+1}-s_i)\left[L_9\frac{\text{log}(\psi(s_{i+1}-s_i))}{\psi(s_{i+1}-s_i)^2}+L_{10}\right]\\
&(L_2L_5\fmax{s\in[0,1]}\|A(s)\|+L_5),
\end{align*}
\normalsize
where we have used \prettyref{Lem:ValueErr} and the property of the eikonal equation that the gradient-norm of a distance function is upper bounded by one or $\left\|\FPPR{\text{dist}(p,\mathcal{E})}{p_2}\right\|\leq1$. To bound the second term, we need to use mean value theorem again. It can be verified that as $d\to0$, the dominating term of $\FPPTR{\text{clog}(d)}{d}$ is $\text{log}(d)/d^3$. Therefore, we can find negative constant $L_{11}$ and positive constant $L_{12}$ such that: 
\small
\begin{align*}
\left|\FPPT{\text{clog}(d)}{d}\right|\leq L_{11}\frac{\text{log}(d)}{d^3}+L_{12}\quad\forall d>0.
\end{align*}
\normalsize
We are now ready to invoke the mean value theorem and bound the second term as:
\small
\begin{align*}
&\left|\FPP{\text{clog}(d)}{d_1}-\FPP{\text{clog}(d)}{d_2}\right|
\left\|\FPP{\text{dist}(p,\mathcal{E})}{p_2}A(s_2)\right\|\\
\leq&\left|\int_{d_1}^{d_2}\FPPT{\text{clog}(d)}{d}dd\right|\fmax{s\in[0,1]}\|A(s)\|\\
%\leq&\int_{d_1}^{d_2}\left[\FPPT{\text{clog}(d)}{d}\right]dd\fmax{s\in[0,1]}\|A(s)\|\\
\leq&\int_{d_1}^{d_2}\left[L_{11}\frac{\text{log}(d)}{d^3}+L_{12}\right]dd\fmax{s\in[0,1]}\|A(s)\|\\
\leq&|d_1-d_2|\fmax{d\in[d_1,d_2]}\left[L_{11}\frac{\text{log}(d)}{d^3}+L_{12}\right]\fmax{s\in[0,1]}\|A(s)\|\\
\leq&L_1L_5(s_{i+1}-s_i)\fmax{d\in[d_1,d_2]}\left[L_{11}\frac{\text{log}(d)}{d^3}+L_{12}\right]\fmax{s\in[0,1]}\|A(s)\|\\
\leq&(s_{i+1}-s_i)\left[L_{11}\frac{\text{log}(\psi(s_{i+1}-s_i))}{\psi(s_{i+1}-s_i)^3}+L_{12}\right]
(L_1L_5\fmax{s\in[0,1]}\|A(s)\|).
\end{align*}
\normalsize
The rest of the argument is identical to \prettyref{Lem:ValueErr}. If $\eta<1/3$, then the leading term of $|\nabla_w\tilde{B}-\nabla_w\hat{B}_{\epsilon_6}|$ is $\epsilon_6^{1-3\eta}\text{log}(\epsilon_6)$, which is diminishing as $\epsilon_6\to0$.
\end{IEEEproof}

\subsection{Finite Termination of Line Search}
It is obvious that if line search does not have finite termination, then Subdivide-Unsafe will be invoked for infinitely many times and $\epsilon_\alpha$ will get arbitrarily small. We first notice that due to the positivity of $\psi$, $\hat{B}_{\epsilon_6}$ is finite:
\begin{corollary}
\label{Cor:FiniteBHat}
If we choose any fixed $\epsilon_6$ and suppose $w$ is some solution generated by an iteration of inexact P-IPM, then $\hat{B}_{\epsilon_6}(w,\mathcal{E})<\infty$.
\end{corollary}
\begin{IEEEproof}
For each point $A(s)w$, we have: $\text{dist}(A(s)w,\mathcal{E})-d_0>\fmin{i=1,\cdots,S-1}\psi(s_{i+1}-s_i)>0$.
\end{IEEEproof}
We can prove by contradiction that if there are infinitely many unsafe B\'ezier curve pieces, than $\hat{B}_{\epsilon_6}$ cannot be finite, which is an indication of finite termination of \prettyref{alg:searchInexact}.
\begin{lemma}
\label{Lem:InfiniteIntegral}
If we choose any fixed $\epsilon_6$ and suppose there are infinitely many calls to \prettyref{alg:subdUnsafe} within \prettyref{alg:pipmInexact}, then $\hat{B}_{\epsilon_6}$ is unbounded.
\end{lemma}
\begin{IEEEproof}
Suppose we have infinitely many calls to \prettyref{alg:subdUnsafe}, then there must be an unsafe B\'ezier curve piece satisfying:
\begin{align*}
s_{i+1}-s_i\leq\frac{1}{2^k}<\epsilon_6,
\end{align*}
for any, sufficiently large, positive integer $k$. Consider a point $p=A(s)w$ on that unsafe curve piece $s\in[s_i,s_{i+1}]$, which is supposed to be updated to $p'=A(s)w'$ after the line search. We can choose $s\in(s_i,s_{i+1})$ such that:
\begin{align*}
&\text{dist}(l_{pp'}',\mathcal{E})<d_0+\psi(s_{i+1}-s_i)\\
\Rightarrow&\text{dist}(p,\mathcal{E})\leq\text{dist}(l_{pp'}',\mathcal{E})+\text{dist}(p,p')\\
\leq&d_0+\psi(s_{i+1}-s_i)+L_2L_5\alpha\|d\|.
\end{align*}
Now consider a small range $[s-\epsilon_5,s+\epsilon_5]\subset(s_i,s_{i+1})$ in which we have:
\begin{align*}
&\text{dist}(A(s')w,\mathcal{E})\leq\text{dist}(p,\mathcal{E})+L_1L_5\epsilon_5\\
\leq&d_0+\psi(s_{i+1}-s_i)+L_2L_5\alpha\|d\|+L_1L_5\epsilon_5,
\end{align*}
where $s'\in[s-\epsilon_5,s+\epsilon_5]$. We can then bound $\hat{B}_{\epsilon_6}$ from below as:
\begin{align*}
\hat{B}_{\epsilon_6}(w,\mathcal{E})\geq&\int_{s-\epsilon_5}^{s+\epsilon_5}
\text{clog}(\text{dist}(A(s')w,\mathcal{E})-d_0)ds'\\
\geq&2\epsilon_5\text{clog}(\psi(s_{i+1}-s_i)+L_2L_5\alpha\|d\|+L_1L_5\epsilon_5)\\
\geq&2\epsilon_5\text{clog}(\frac{\epsilon_s}{2^{\eta k}}+L_2L_5\epsilon_\alpha\|d\|+L_1L_5\epsilon_5).
\end{align*}
Taking $k\to\infty$, $\epsilon_5\to0$, $\epsilon_\alpha\to0$, and we have $\hat{B}_{\epsilon_6}(w)\to\infty$, contradicting \prettyref{Cor:FiniteBHat}.
\end{IEEEproof}
The finite termination of \prettyref{alg:searchInexact} is a direct corollary of the two results above:
\begin{corollary}
\prettyref{alg:searchInexact} has finite termination.
\end{corollary}
\begin{IEEEproof}
If \prettyref{alg:searchInexact} does not have finite termination, then $\hat{B}_{\epsilon_6}$ is unbounded by \prettyref{Lem:InfiniteIntegral}, contradicting \prettyref{Cor:FiniteBHat}.
\end{IEEEproof}

\subsection{Finite Termination of Outer Loop}
Similar to the analysis of exact P-IPM, we show finite termination for gradient descend method (i.e. $\text{SPD}(\nabla_{w,T}^2 O)=I$), instead of Newton's method, for simplicity. Unlike exact P-IPM, however, the inexact line search \prettyref{alg:searchInexact} is not guaranteed to reduce function value, because the function value might increase due to \prettyref{alg:subdUnsafe}. Instead, we analyze the total number of subdivisions throughout \prettyref{alg:pipmInexact} in two cases.

\textbf{Case I:} If \prettyref{alg:pipmInexact} uses finitely many unsafe subdivisions, then we can find a last outer iteration calling \prettyref{alg:subdUnsafe}. Afterwards, we are in a similar situation to \prettyref{Prop:ExactTerminate}.
\begin{lemma}
\label{Lem:FiniteInexact}
Assuming \prettyref{Ass:A1}, if \prettyref{alg:pipmInexact} makes finitely many calls to \prettyref{alg:subdUnsafe}, then \prettyref{alg:pipmInexact} terminates after finitely many iterations.
\end{lemma}
\begin{IEEEproof}
If there are only finitely many calls to \prettyref{alg:subdUnsafe}, then we can denote $w$ as the solution generated by the outer iteration that makes the last call to \prettyref{alg:subdUnsafe}. After this iteration, $\epsilon_a$ takes some positive value. From this iteration onward, $\epsilon_a$ will never change and $\epsilon_a$ is always safe, because there will be another call to \prettyref{alg:subdUnsafe} otherwise. Therefore, we have finite termination by \cite[Proposition~1.2.4]{bertsekas1997nonlinear}.
\end{IEEEproof}

\textbf{Case II:} If \prettyref{alg:pipmInexact} uses infinitely many unsafe subdivisions, then $\hat{B}_{\epsilon_6}$ increases but $\tilde{B}$ decreases during an update from $w$ to $w'$. But \prettyref{Lem:ValueErr} and \prettyref{Lem:GradErr} shows that $\hat{B}_{\epsilon_6}$ and $\tilde{B}$ can be arbitrarily close, from which we can derive a contradiction.
\begin{lemma}
\label{Lem:InfiniteInexact}
Assuming \prettyref{Ass:A1}, \prettyref{alg:pipmInexact} makes finitely many calls to \prettyref{alg:subdUnsafe}.
\end{lemma}
\begin{IEEEproof}
Let's use the following abbreviations of the objective functions:
\begin{align*}
\tilde{f}(w)\triangleq&O(w,T) + \lambda \tilde{B}(w,\mathcal{E})\\
\hat{f}_{\epsilon_6}(w)\triangleq&O(w,T) + \lambda \hat{B}_{\epsilon_6}(w,\mathcal{E}).
\end{align*}
If there are infinitely many calls to \prettyref{alg:subdUnsafe}, then $\|\nabla_w\tilde{f}\|>\epsilon_g$ because the outer loop would terminate immediately otherwise. Consider a update from $w$ to $w'$ which satisfies the Wolfe's condition, we have:
\begin{align*}
\tilde{f}(w')\leq\tilde{f}(w)-c\alpha\|\nabla_w\tilde{f}\|^2
\leq \tilde{f}(w)-c\alpha\|\nabla_w\tilde{f}\|\epsilon_g.
\end{align*}
We can bound the corresponding change in $\hat{f}_{\epsilon_6}$ using \prettyref{Lem:GradErr} as follows:
\footnotesize
\begin{align*}
\hat{f}_{\epsilon_6}(w')
=&\hat{f}_{\epsilon_6}(w)+\int_{w}^{w'}
\left<\nabla_w\hat{f}_{\epsilon_6}(\omega),d\omega\right>\\
=&\hat{f}_{\epsilon_6}(w)+\int_{w}^{w'}
\left<\nabla_w\hat{f}_{\epsilon_6}(\omega)-\nabla_w\tilde{f}(\omega)+\nabla_w\tilde{f}(\omega),d\omega\right>\\
\leq&\hat{f}_{\epsilon_6}(w)+\int_{w}^{w'}
\|\nabla_w\hat{f}_{\epsilon_6}(\omega)-\nabla_w\tilde{f}(\omega)\|\|d\omega\|+
\tilde{f}(w')-\tilde{f}(w)\\
\leq&\hat{f}_{\epsilon_6}(w)+
\mathcal{O}(\epsilon_6^{1-3\eta}\text{log}(\epsilon_6))\|w'-w\|-c\alpha\|\nabla_w\tilde{f}\|\epsilon_g\\
%=&\hat{f}_{\epsilon_6}(w)+
%\mathcal{O}(\epsilon_6^{1-3\eta}\text{log}(\epsilon_6))\alpha\|\nabla_w\tilde{f}\|-c\alpha\|\nabla_w\tilde{f}\|\epsilon_g\\
=&\hat{f}_{\epsilon_6}(w)+\alpha\|\nabla_w\tilde{f}\|
\left[\mathcal{O}(\epsilon_6^{1-3\eta}\text{log}(\epsilon_6))-c\epsilon_g\right],
\end{align*}
\normalsize
where we can choose fixed, sufficiently small $\epsilon_6$ so that:
\begin{align*}
\hat{f}_{\epsilon_6}(w')\leq\hat{f}_{\epsilon_6}(w).
\end{align*}
Moreover, by \prettyref{Lem:BHatInvariance}, $\hat{f}$ is invariant to subdivision after finitely many calls to \prettyref{alg:subdUnsafe}. We conclude that, after finitely many iterations, $\hat{f}_{\epsilon_6}$ monotonically decreases, contradicting \prettyref{Lem:InfiniteIntegral}.
\end{IEEEproof}

\begin{IEEEproof}[Proof of proposition \prettyref{Props:pipmInexact}]
The outer loop of \prettyref{alg:pipmInexact} has finite termination by combining \prettyref{Lem:FiniteInexact} and \prettyref{Lem:InfiniteInexact}. 
\end{IEEEproof}

%% file: 00-main.bbl
\begin{thebibliography}{10}
\providecommand{\url}[1]{#1}
\csname url@rmstyle\endcsname
\providecommand{\newblock}{\relax}
\providecommand{\bibinfo}[2]{#2}
\providecommand\BIBentrySTDinterwordspacing{\spaceskip=0pt\relax}
\providecommand\BIBentryALTinterwordstretchfactor{4}
\providecommand\BIBentryALTinterwordspacing{\spaceskip=\fontdimen2\font plus
\BIBentryALTinterwordstretchfactor\fontdimen3\font minus
  \fontdimen4\font\relax}
\providecommand\BIBforeignlanguage[2]{{%
\expandafter\ifx\csname l@#1\endcsname\relax
\typeout{** WARNING: IEEEtran.bst: No hyphenation pattern has been}%
\typeout{** loaded for the language `#1'. Using the pattern for}%
\typeout{** the default language instead.}%
\else
\language=\csname l@#1\endcsname
\fi
#2}}

\bibitem{berry1997uniqueness}
T.~G. Berry and R.~R. Patterson, ``The uniqueness of b{\'e}zier control
  points,'' \emph{Computer Aided Geometric Design}, vol.~14, no.~9, pp.
  877--879, 1997.

\bibitem{bertsekas1997nonlinear}
D.~P. Bertsekas, ``Nonlinear programming,'' \emph{Journal of the Operational
  Research Society}, vol.~48, no.~3, pp. 334--334, 1997.

\bibitem{10.1145/2185520.2185592}
\BIBentryALTinterwordspacing
T.~Brochu, E.~Edwards, and R.~Bridson, ``Efficient geometrically exact
  continuous collision detection,'' \emph{ACM Trans. Graph.}, vol.~31, no.~4,
  July 2012. [Online]. Available: \url{https://doi.org/10.1145/2185520.2185592}
\BIBentrySTDinterwordspacing

\bibitem{caratheodory1911variabilitatsbereich}
C.~Carath{\'e}odory, ``{\"U}ber den variabilit{\"a}tsbereich der
  fourier’schen konstanten von positiven harmonischen funktionen,''
  \emph{Rendiconti Del Circolo Matematico di Palermo (1884-1940)}, vol.~32,
  no.~1, pp. 193--217, 1911.

\bibitem{curtis2012sequential}
F.~E. Curtis and M.~L. Overton, ``A sequential quadratic programming algorithm
  for nonconvex, nonsmooth constrained optimization,'' \emph{SIAM Journal on
  Optimization}, vol.~22, no.~2, pp. 474--500, 2012.

\bibitem{deits2015efficient}
R.~Deits and R.~Tedrake, ``Efficient mixed-integer planning for uavs in
  cluttered environments,'' in \emph{2015 IEEE international conference on
  robotics and automation (ICRA)}.\hskip 1em plus 0.5em minus 0.4em\relax IEEE,
  2015, pp. 42--49.

\bibitem{sketchfab}
\BIBentryALTinterwordspacing
A.~Denoyel, C.~Pinson, and P.-A. Passet, \emph{SketchFab}, 2020 (accessed
  October 25, 2020). [Online]. Available: \url{https://sketchfab.com/}
\BIBentrySTDinterwordspacing

\bibitem{di2015energy}
C.~Di~Franco and G.~Buttazzo, ``Energy-aware coverage path planning of uavs,''
  in \emph{2015 IEEE international conference on autonomous robot systems and
  competitions}.\hskip 1em plus 0.5em minus 0.4em\relax IEEE, 2015, pp.
  111--117.

\bibitem{ding2018trajectory}
W.~Ding, W.~Gao, K.~Wang, and S.~Shen, ``Trajectory replanning for quadrotors
  using kinodynamic search and elastic optimization,'' in \emph{2018 IEEE
  International Conference on Robotics and Automation (ICRA)}.\hskip 1em plus
  0.5em minus 0.4em\relax IEEE, 2018, pp. 7595--7602.

\bibitem{10.5555/557058}
G.~E. Farin and D.~Hansford, \emph{The Essentials of CAGD}, 1st~ed.\hskip 1em
  plus 0.5em minus 0.4em\relax USA: A. K. Peters, Ltd., 2000.

\bibitem{8206214}
F.~{Gao}, Y.~{Lin}, and S.~{Shen}, ``Gradient-based online safe trajectory
  generation for quadrotor flight in complex environments,'' in \emph{2017
  IEEE/RSJ International Conference on Intelligent Robots and Systems (IROS)},
  2017, pp. 3681--3688.

\bibitem{8462878}
F.~{Gao}, W.~{Wu}, Y.~{Lin}, and S.~{Shen}, ``Online safe trajectory generation
  for quadrotors using fast marching method and bernstein basis polynomial,''
  in \emph{2018 IEEE International Conference on Robotics and Automation
  (ICRA)}, 2018, pp. 344--351.

\bibitem{gravesen1997adaptive}
J.~Gravesen, ``Adaptive subdivision and the length and energy of b{\'e}zier
  curves,'' \emph{Computational Geometry}, vol.~8, no.~1, pp. 13--31, 1997.

\bibitem{grundel2004formulation}
D.~Grundel and D.~Jeffcoat, ``Formulation and solution of the target visitation
  problem,'' in \emph{AIAA 1st Intelligent Systems Technical Conference}, 2004,
  p. 6212.

\bibitem{guerrero2013uav}
J.~A. Guerrero and Y.~Bestaoui, ``Uav path planning for structure inspection in
  windy environments,'' \emph{Journal of Intelligent \& Robotic Systems},
  vol.~69, no. 1-4, pp. 297--311, 2013.

\bibitem{hauser2018semi}
K.~Hauser, ``Semi-infinite programming for trajectory optimization with
  nonconvex obstacles,'' in \emph{International Workshop on the Algorithmic
  Foundations of Robotics}.\hskip 1em plus 0.5em minus 0.4em\relax Springer,
  2018, pp. 565--580.

\bibitem{kalakrishnan2011stomp}
M.~Kalakrishnan, S.~Chitta, E.~Theodorou, P.~Pastor, and S.~Schaal, ``Stomp:
  Stochastic trajectory optimization for motion planning,'' in \emph{2011 IEEE
  international conference on robotics and automation}.\hskip 1em plus 0.5em
  minus 0.4em\relax IEEE, 2011, pp. 4569--4574.

\bibitem{6425970}
S.~{Kim}, K.~{Sreenath}, S.~{Bhattacharya}, and V.~{Kumar}, ``Optimal
  trajectory generation under homology class constraints,'' in \emph{2012 IEEE
  51st IEEE Conference on Decision and Control (CDC)}, 2012, pp. 3157--3164.

\bibitem{10.1145/3386569.3392425}
\BIBentryALTinterwordspacing
M.~Li, Z.~Ferguson, T.~Schneider, T.~Langlois, D.~Zorin, D.~Panozzo, C.~Jiang,
  and D.~M. Kaufman, ``Incremental potential contact: Intersection-and
  inversion-free, large-deformation dynamics,'' \emph{ACM Trans. Graph.},
  vol.~39, no.~4, July 2020. [Online]. Available:
  \url{https://doi.org/10.1145/3386569.3392425}
\BIBentrySTDinterwordspacing

\bibitem{li2016asymptotically}
Y.~Li, Z.~Littlefield, and K.~E. Bekris, ``Asymptotically optimal
  sampling-based kinodynamic planning,'' \emph{The International Journal of
  Robotics Research}, vol.~35, no.~5, pp. 528--564, 2016.

\bibitem{likhachev2005anytime}
M.~Likhachev, D.~Ferguson, G.~Gordon, A.~Stentz, and S.~Thrun, ``Anytime
  dynamic a*: An anytime, replanning algorithm,'' in \emph{Proceedings of the
  Fifteenth International Conference on International Conference on Automated
  Planning and Scheduling}, ser. ICAPS'05.\hskip 1em plus 0.5em minus
  0.4em\relax AAAI Press, 2005, p. 262–271.

\bibitem{liu2017search}
S.~Liu, N.~Atanasov, K.~Mohta, and V.~Kumar, ``Search-based motion planning for
  quadrotors using linear quadratic minimum time control,'' in \emph{2017
  IEEE/RSJ international conference on intelligent robots and systems
  (IROS)}.\hskip 1em plus 0.5em minus 0.4em\relax IEEE, 2017, pp. 2872--2879.

\bibitem{mechali2019rectified}
O.~Mechali, L.~Xu, M.~Wei, I.~Benkhaddra, F.~Guo, and A.~Senouci, ``A rectified
  rrt* with efficient obstacles avoidance method for uav in 3d environment,''
  in \emph{2019 IEEE 9th Annual International Conference on CYBER Technology in
  Automation, Control, and Intelligent Systems (CYBER)}.\hskip 1em plus 0.5em
  minus 0.4em\relax IEEE, 2019, pp. 480--485.

\bibitem{5980409}
D.~{Mellinger} and V.~{Kumar}, ``Minimum snap trajectory generation and control
  for quadrotors,'' in \emph{2011 IEEE International Conference on Robotics and
  Automation}, 2011, pp. 2520--2525.

\bibitem{5569026}
N.~{Michael}, D.~{Mellinger}, Q.~{Lindsey}, and V.~{Kumar}, ``The grasp
  multiple micro-uav testbed,'' \emph{IEEE Robotics Automation Magazine},
  vol.~17, no.~3, pp. 56--65, 2010.

\bibitem{montanari2017improving}
M.~Montanari, N.~Petrinic, and E.~Barbieri, ``Improving the gjk algorithm for
  faster and more reliable distance queries between convex objects,'' \emph{ACM
  Transactions on Graphics (TOG)}, vol.~36, no.~3, pp. 1--17, 2017.

\bibitem{mordukhovich2013constraint}
B.~Mordukhovich and T.~Nghia, ``Constraint qualifications and optimality
  conditions for nonconvex semi-infinite and infinite programs,''
  \emph{Mathematical Programming}, vol. 139, no. 1-2, pp. 271--300, 2013.

\bibitem{ruiqi2021}
R.~Ni, T.~Schneider, D.~Panozzo, Z.~Pan, and X.~Gao, ``Robust \& asymptotically
  locally optimal uav-trajectory generation based on spline subdivision,''
  \emph{International Conference on Robotics and Automation (ICRA)}, 2021.

\bibitem{park2012itomp}
C.~Park, J.~Pan, and D.~Manocha, ``Itomp: Incremental trajectory optimization
  for real-time replanning in dynamic environments,'' in \emph{Twenty-Second
  International Conference on Automated Planning and Scheduling}, 2012.

\bibitem{pham2014general}
Q.-C. Pham, ``A general, fast, and robust implementation of the time-optimal
  path parameterization algorithm,'' \emph{IEEE Transactions on Robotics},
  vol.~30, no.~6, pp. 1533--1540, 2014.

\bibitem{10.5555/581820}
H.~Prautzsch, W.~Boehm, and M.~Paluszny, \emph{Bezier and B-Spline
  Techniques}.\hskip 1em plus 0.5em minus 0.4em\relax Berlin, Heidelberg:
  Springer-Verlag, 2002.

\bibitem{richter2016polynomial}
C.~Richter, A.~Bry, and N.~Roy, ``Polynomial trajectory planning for aggressive
  quadrotor flight in dense indoor environments,'' in \emph{Robotics
  Research}.\hskip 1em plus 0.5em minus 0.4em\relax Springer, 2016, pp.
  649--666.

\bibitem{rockafellar2015convex}
R.~T. Rockafellar, \emph{Convex analysis}.\hskip 1em plus 0.5em minus
  0.4em\relax Princeton university press, 2015.

\bibitem{schulman2014motion}
J.~Schulman, Y.~Duan, J.~Ho, A.~Lee, I.~Awwal, H.~Bradlow, J.~Pan, S.~Patil,
  K.~Goldberg, and P.~Abbeel, ``Motion planning with sequential convex
  optimization and convex collision checking,'' \emph{The International Journal
  of Robotics Research}, vol.~33, no.~9, pp. 1251--1270, 2014.

\bibitem{slegers2006nonlinear}
N.~Slegers, J.~Kyle, and M.~Costello, ``Nonlinear model predictive control
  technique for unmanned air vehicles,'' \emph{Journal of guidance, control,
  and dynamics}, vol.~29, no.~5, pp. 1179--1188, 2006.

\bibitem{sun2020fast}
W.~Sun, G.~Tang, and K.~Hauser, ``Fast uav trajectory optimization using
  bilevel optimization with analytical gradients,'' in \emph{2020 American
  Control Conference (ACC)}.\hskip 1em plus 0.5em minus 0.4em\relax IEEE, 2020,
  pp. 82--87.

\bibitem{9196789}
G.~{Tang}, W.~{Sun}, and K.~{Hauser}, ``Enhancing bilevel optimization for uav
  time-optimal trajectory using a duality gap approach,'' in \emph{2020 IEEE
  International Conference on Robotics and Automation (ICRA)}, 2020, pp.
  2515--2521.

\bibitem{tang2014fast}
M.~Tang, R.~Tong, Z.~Wang, and D.~Manocha, ``Fast and exact continuous
  collision detection with bernstein sign classification,'' \emph{ACM
  Transactions on Graphics (TOG)}, vol.~33, no.~6, pp. 1--8, 2014.

\bibitem{von1993numerical}
O.~Von~Stryk, ``Numerical solution of optimal control problems by direct
  collocation,'' in \emph{Optimal control}.\hskip 1em plus 0.5em minus
  0.4em\relax Springer, 1993, pp. 129--143.

\bibitem{9121729}
Z.~{Wang}, X.~{Zhou}, C.~{Xu}, J.~{Chu}, and F.~{Gao}, ``Alternating
  minimization based trajectory generation for quadrotor aggressive flight,''
  \emph{IEEE Robotics and Automation Letters}, vol.~5, no.~3, pp. 4836--4843,
  2020.

\bibitem{webb2012kinodynamic}
D.~J. Webb and J.~van~den Berg, ``Kinodynamic rrt*: Optimal motion planning for
  systems with linear differential constraints,'' \emph{arXiv}, pp.
  arXiv--1205, 2012.

\bibitem{webb2013kinodynamic}
D.~J. Webb and J.~Van Den~Berg, ``Kinodynamic rrt*: Asymptotically optimal
  motion planning for robots with linear dynamics,'' in \emph{2013 IEEE
  International Conference on Robotics and Automation}.\hskip 1em plus 0.5em
  minus 0.4em\relax IEEE, 2013, pp. 5054--5061.

\bibitem{williams2016aggressive}
G.~Williams, P.~Drews, B.~Goldfain, J.~M. Rehg, and E.~A. Theodorou,
  ``Aggressive driving with model predictive path integral control,'' in
  \emph{2016 IEEE International Conference on Robotics and Automation
  (ICRA)}.\hskip 1em plus 0.5em minus 0.4em\relax IEEE, 2016, pp. 1433--1440.

\bibitem{8758904}
B.~{Zhou}, F.~{Gao}, L.~{Wang}, C.~{Liu}, and S.~{Shen}, ``Robust and efficient
  quadrotor trajectory generation for fast autonomous flight,'' \emph{IEEE
  Robotics and Automation Letters}, vol.~4, no.~4, pp. 3529--3536, 2019.

\bibitem{zucker2013chomp}
M.~Zucker, N.~Ratliff, A.~D. Dragan, M.~Pivtoraiko, M.~Klingensmith, C.~M.
  Dellin, J.~A. Bagnell, and S.~S. Srinivasa, ``Chomp: Covariant hamiltonian
  optimization for motion planning,'' \emph{The International Journal of
  Robotics Research}, vol.~32, no. 9-10, pp. 1164--1193, 2013.

\end{thebibliography}
